\documentclass[PHM, 2024]{PHMSociety}

\usepackage{graphicx}
\usepackage{amsmath}
\usepackage{multirow}
\usepackage{subcaption}
\usepackage{url}
\usepackage{ragged2e}
\begin{document}

\title{Adversarial Attacks and Defenses in Multivariate Time-Series Forecasting for Smart and Connected Infrastructures}

\author{%
	Pooja Krishan \authorNumber{1}, Rohan Mohapatra \authorNumber{2}, Sanchari Das \authorNumber{3} and Saptarshi Sengupta \authorNumber{4}
}

    

\address{
	\affiliation{{1,2,4}}{Department of Computer Science, San Jos\'e State University, San Jos\'e, CA, 95192, USA}
     \affiliation{{3}}{Department of Computer Science, University of Denver, Denver, CO, 80210, USA}{ 
		{\email{pooja.krishan@sjsu.edu}}\\ 
		{\email{rohan.mohapatra@sjsu.edu}}
  \\        {\email{sanchari.das@du.edu}}\\
		{\email{saptarshi.sengupta@sjsu.edu}}
		} 
}

\maketitle

\phmLicenseFootnote{Pooja Krishan}

\begin{abstract}
The emergence of deep learning models has revolutionized various industries over the last decade, leading to a surge in connected devices and infrastructures. However, these models can be tricked into making incorrect predictions with high confidence, leading to disastrous failures and security concerns. To this end, we explore the impact of adversarial attacks on multivariate time-series forecasting and investigate methods to counter them. Specifically, we employ untargeted white-box attacks, namely the Fast Gradient Sign Method (FGSM) and the Basic Iterative Method (BIM), to poison the inputs to the training process, effectively misleading the model. We also illustrate the subtle modifications to the inputs after the attack, which makes detecting the attack using the naked eye quite difficult. Having demonstrated the feasibility of these attacks, we develop robust models through adversarial training and model hardening. We are among the first to showcase the transferability of these attacks and defenses by extrapolating our work from the benchmark electricity data to a larger, 10-year real-world data used for predicting the time-to-failure of hard disks. Our experimental results confirm that the attacks and defenses achieve the desired security thresholds, leading to a 72.41\% and 94.81\% decrease in RMSE for the electricity and hard disk datasets respectively after implementing the adversarial defenses.
\end{abstract}

\section{Introduction}\label{sec:intro}
A time-series records a series of metrics over regular intervals of time as a sequence of values. 

Time-series forecasting is the task of estimating the output at a certain time step, given the previous values. It is used in a variety of domains such as finance \cite{ts-finance}, power consumption prediction \cite{ts-power-consumption}, health prediction of equipment \cite{ts-health-prediction}, healthcare \cite{ts-healthcare}, and weather forecasting \cite{ts-weather-forecasting}. The widespread use of sensors and actuators has resulted in a proliferation of data, leading to the shift from traditional time-series forecasting methods to deep learning architectures \cite{deep-learning-superior}, which are more capable of gleaning insights and identifying long-term trends from the data. However, it is a double-edged sword as deep learning models can be easily compromised by attacks, causing the models to produce incorrect forecasts based on manipulated input data. This gullible nature of deep learning models to attacks paves the way for catastrophic failures in safety-critical applications and leads to the wastage of valuable resources, time, money, and productivity \cite{dl-model-threat}. This opens up a new area of research to develop models resistant to these types of attacks.

Adversarial attacks on deep learning models are classified into white-box or black-box attacks, and targeted or untargeted attacks depending on the ease of access, and the attacker's goal respectively. In white-box attacks, the attacker knows sensitive model-specific information such as inputs, targets, and gradients \cite{melis2021explaining}. Conversely, in black-box attacks, the model is viewed as an oracle that outputs values given input data and the attack is crafted based on observed model behavior \cite{oh2019towards,tsingenopoulos2019autoattacker}. In targeted attacks, the adversary tries to not only delude the model but also prompts it to produce an output from a particular distribution \cite{fursov2021adversarial} whereas in untargeted attacks the attacker intends to trigger the model to generate incorrect outputs belonging to any distribution \cite{miller2020adversarial,lin2021ml}.

Adversarial defense involves training the deep learning network with augmented data capturing the noise distribution that the attacker plans on using, or modifying the model architecture to enhance the robustness of the model \cite{tariq2020review,akhtar2021advances}.

In this study, we train a Long Short-Term Memory (LSTM) model \cite{lstm} on a toy dataset that is used to predict future power consumption, given the past values to avoid under-utilization or over-utilization of resources. After conducting numerous training experiments, we select the best-performing LSTM model. We then subject this model to two untargeted white box attacks – the Fast Gradient Sign Method (FGSM), and the Basic Iterative Method (BIM), causing it to learn the underlying distribution incorrectly and leading to an increased error rate. Once we demonstrate the impact and the ease of the attacks, we move on to implement adversarial defense using both a data augmentation-based approach, and a layer-wise hardening of the neural network weights enabling the model to learn over the poisoned noise distribution in addition to the actual training samples. We then repeat the above set of experiments on a large-scale hard disk drive dataset that predicts the Remaining Useful Life (RUL) to show the transferability of the attacks and defense schemes proposed. 

Our key contributions are:
\begin{enumerate}
    \item \textbf{Efficient training:} We run multiple experiments to identify the best deep learning model.
    \item \textbf{Effective attacks:} We successfully demonstrate the impact of the adversarial attacks in all the datasets used.
    \item \textbf{Risk mitigation:} We perform different adversarial defenses to develop models resilient to attacks.
    \item \textbf{Imperceptibility of perturbation:} We visualize the indiscernible changes to the input after the attack.
    \item \textbf{Widespread applicability:} We use two datasets to prove the efficacy and transferability of the attacks and defenses.
\end{enumerate}

The paper is outlined as follows: In Section~\ref{sec:lit-review}, we will go over previous literature and identify the opportunities in this domain. In Section~\ref{sec:exp-setup} we will walk through the preprocessing of the datasets used in the experiments providing an insight into the deep learning models and training parameters used in Section~\ref{sec:model_training}. In Section~\ref{sec:overview} we summarize the overall attacks and defense schemes used in this paper. We summarize the results in Section~\ref{sec:results}. In Section~\ref{sec:future_work} we outline the future directions, and finally conclude in Section~\ref{sec:conclusion}.

\section{Related Work}\label{sec:lit-review}

\cite{sensors-big-data} outline the interdependence between Internet of Things (IoT) and Big Data. With the proliferation of sensors that record and share information between devices on the Internet, there is no shortage of data, and developing deep learning models to predict future sensor values has become easy. This has led to a shift from traditional methods of time-series forecasting using model-driven methods to a data-driven method involving multiple deep learning models for time-series forecasting problems \cite{shift}. \cite{ts1} found that LSTM models are better at predicting electricity consumed, given exogenous attributes such as temperature, humidity, wind speed, etc. \cite{ts2} prove that using a Gated Recurrent Unit (GRU) \cite{cho} to predict power consumed in New South Wales in Australia led to much better forecasting results than using traditional models. \cite{ts3} use a dilated temporal CNN to capture load consumption using multiple synthetic and real-world datasets accurately. \cite{ts4} prove that a Bidirectional Recurrent Neural Network (BRNN) is better at forecasting traffic flows using the GPS data collected from the Hohhot Bus Corporation than LSTM or GRU models. \cite{ts5} showcase the advantages of using deep networks in comparison to shallow ones while predicting stock prices. \cite{ts9} compare conventional machine learning models to deep learning ones which use a combination of wavelet decomposition and LSTMs in stock market predictions. \cite{ts6} use an LSTM working through the attention mechanism to predict the RUL of aircraft engines. \cite{ts7} have used a neural network model combining LSTM and Convolutional Neural Network (CNN) \cite{cnn} to predict the RUL of aircraft engines. \cite{ts8} show the advantages of using a deep learning model to predict the RUL of spinning parts.

Despite these advances, deep learning models remain vulnerable to adversarial attacks. \cite{szegedy} reported vulnerabilities in deep learning models in image recognition, where Convolutional Neural Networks (CNNs) can be manipulated by injecting minute modifications into the data, thereby leading the network to miscalculate the input with high conviction. \cite{attack-survey} have presented a survey of all types of antagonistic manipulations and their impacts on image recognition such as self-driving cars \cite{road-sign}, robotic vision \cite{robot-vision}, cyberspace attacks \cite{cyber-space}, etc. With the development of multiple threat models to perform adversarial attacks, it is no surprise that a lot of effort went into developing adversarial defense strategies for these attacks in image recognition. \cite{attack-defense-survey} summarize the adversarial attacks and defense mechanisms developed in computer vision in recent years. 

\begin{figure}[!htb]
  \centering
  \includegraphics[width=1\linewidth]{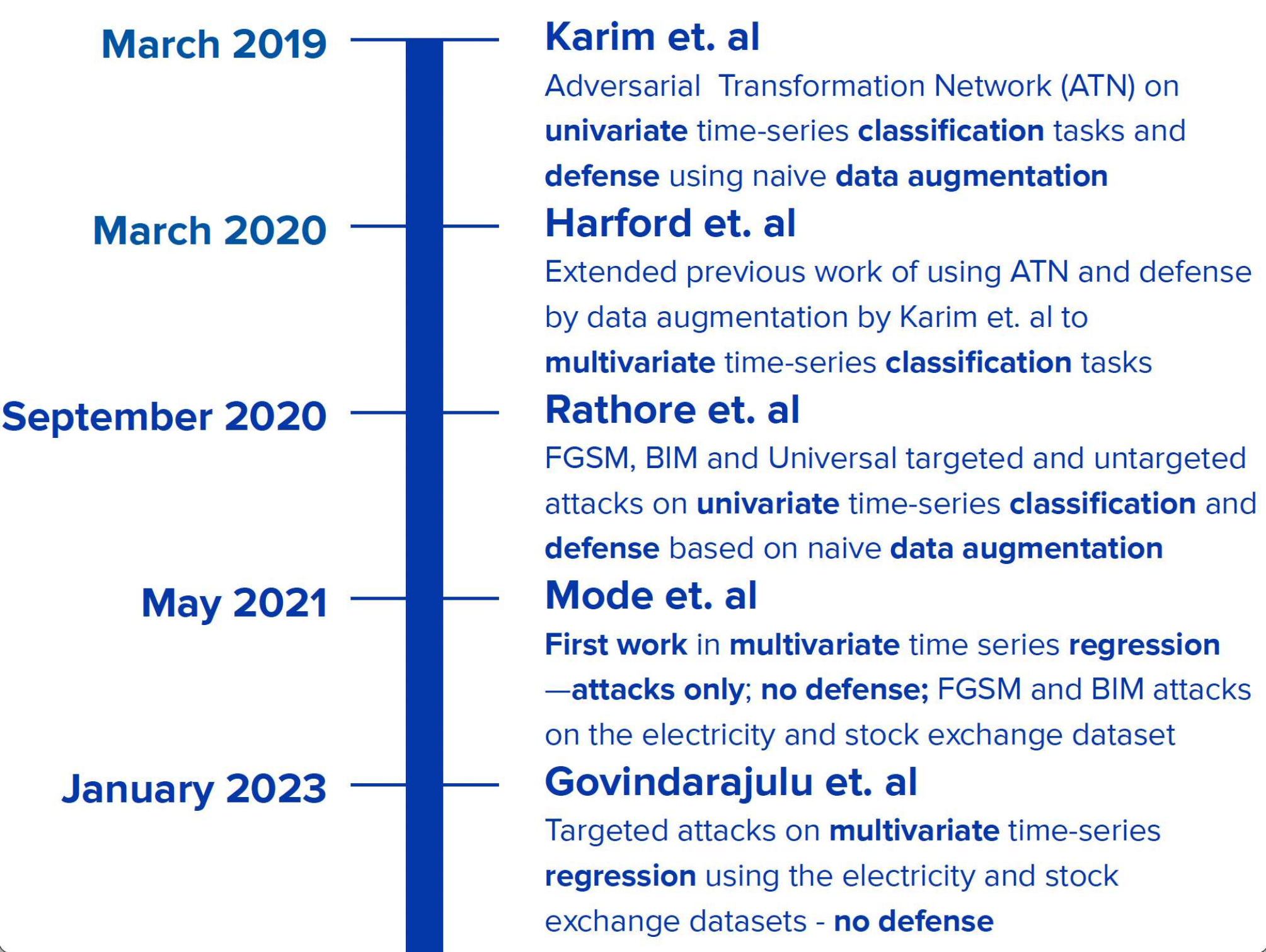}
  \caption{Overview of previous attacks and defenses work in the time-series forecasting domain.}
  \label{fig:lit-review}
\end{figure}

In recent times, adversarial attacks and defense strategies used in time-series analysis have garnered the interest of researchers. \cite{karim} have proposed an Adversarial Transformation Network (ATN) to attack univariate temporal sequential data models used in classification tasks. They have also successfully defended against these attacks using a naive method of data augmentation. \cite{harford} have extended the previous work to multivariate time-series classification problems. \cite{rathore} have demonstrated the effect of FGSM, BIM, and Universal targeted and untargeted attacks on univariate time-series classification tasks, and defense based on the primitive data augmentation technique. \cite{mode} have proposed traditional attacks on multivariate time-series regression datasets such as the Google Stock dataset from Nasdaq and the Electricity dataset \cite{uci-dataset}. They show the vulnerability of CNNs, LSTMs, and GRUs. \cite{govindarajulu} have carried out targeted attacks based on amplitude, direction, and temporal components of the model and showed their effectiveness using statistical tests on the Google stock exchange and Electricity datasets. This is visually represented in \figurename~\ref{fig:lit-review}.

Our extensive review of the literature opened up a realm of opportunities for improvement of current state-of-the-art methods. We observed the following:
\begin{enumerate}
    \item All previous work has been done only on toy datasets such as the Electricity and stock exchange datasets
    \item Adversarial defense has not been performed on the multivariate time-series datasets
    \item The literature also lacked in demonstrating the imperceptible nature of the adversarial attacks on visualization of training sample inputs 
\end{enumerate}

We were motivated to address the research gaps by:
\begin{enumerate}
    \item Extending our experiments on the toy Electricity dataset to a real-world dataset predicting the Remaining Useful Life (RUL) of Hard Disk Drives (HDDs)
    \item Performing adversarial defenses:
    \begin{enumerate}
        \item Using data augmentation-based adversarial training
        \item Using model hardening techniques that perturb the gradients of the model during the training process to successfully defend against adversarial attacks
    \end{enumerate}
    \item Demonstrating the indiscernible nature of the attacks to the naked-eye by visualizing the training sample inputs to the machine learning models, after performing the adversarial attacks
\end{enumerate}

\section{Experimental Setup}\label{sec:exp-setup}
We first carry out our experiments on a smaller dataset which is used to predict the power consumed sometime in the future, given past readings. Once we have proved the success of the attacks and defense techniques in the electricity dataset, we repeat the experiments on a substantially larger dataset which is used to predict the RUL of Hard Disk Drives (HDDs) to prove the ease of transferability of these attacks and defenses. We describe the datasets used in this research and the preparation steps done to make the data more viable for ingestion by the deep learning models in this section. 

\subsection{Individual Household Power Consumption Dataset}
Power consumption prediction is a vital task to estimate the amount of power that has to be supplied to various locations at any given time. It has a direct bearing on the environment and helps to cut costs. Motivated by the applications of power consumption prediction and to demonstrate the effects of antagonistic manipulation and fortification on a multivariate time-series dataset, we chose the Electricity dataset from the UCI repository \cite{uci-dataset} in this research. 

It consists of over 2 million rows and 9 columns sampled by minute in a household for 4 years from the end of 2006 to the end of 2010. \textit{global active power} represents the active power consumed in kW and \textit{global reactive power} represents the reactive power consumed in kW, \textit{voltage} and \textit{global intensity} represents the mean voltage in volts and mean current in amperes respectively. \textit{sub metering 1}, \textit{sub metering 2}, and \textit{sub metering 3} is the active energy readings from the kitchen, laundry room, and electric water heating and air conditioning equipment respectively. The unit of all sub-metering readings is in watt-hour. The active term refers to the total actual power consumed whereas the reactive term corresponds to the unused power in the transmission wires.

\begin{figure}[!htb]
  \centering
  \includegraphics[width=1\linewidth]{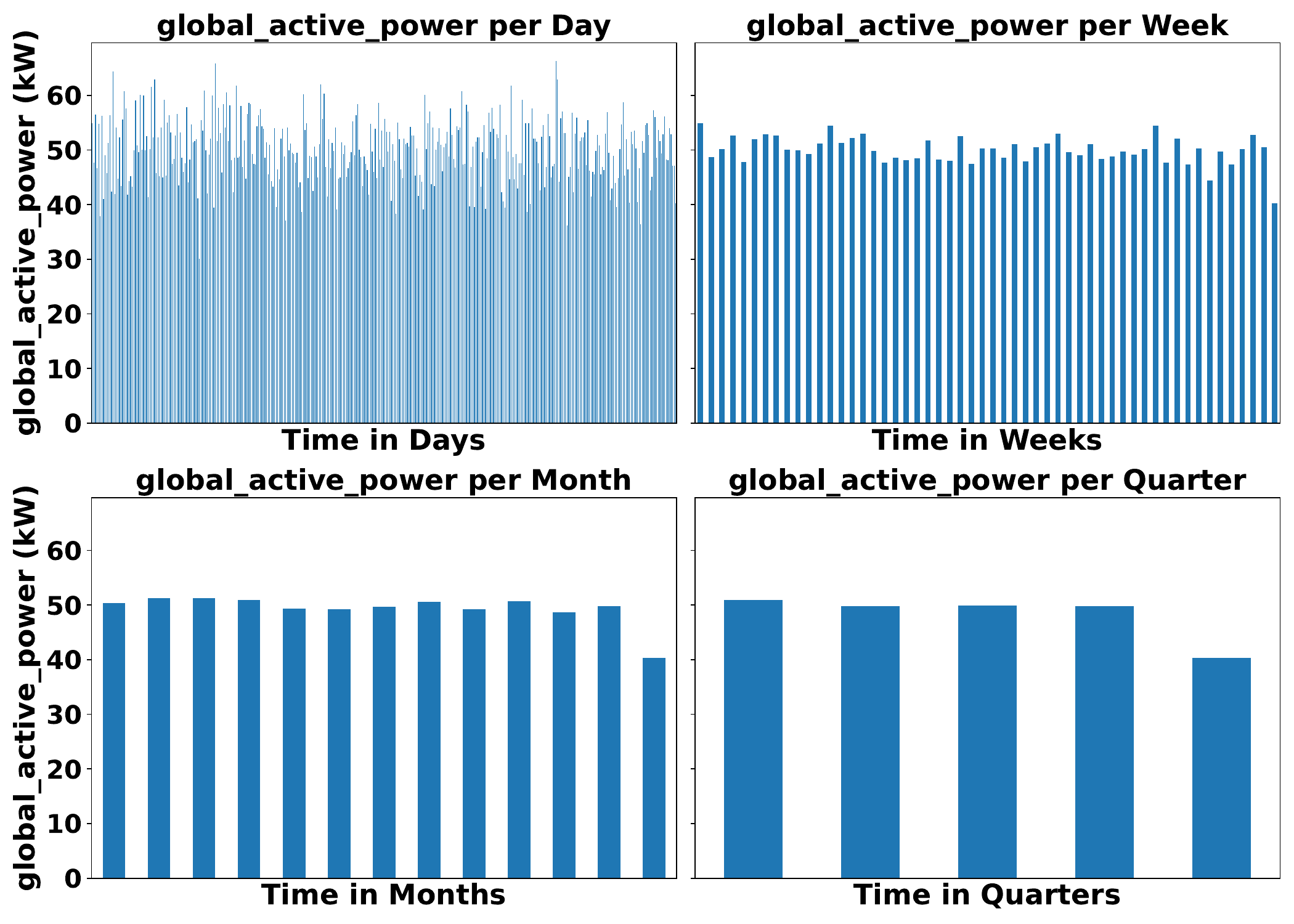}
  \caption{Periodicity decreases as time interval increases.}
  \label{fig:periodicity}
\end{figure}

During preprocessing of the dataset, we treated the missing values (\textit{?}) as null values (\textit{NaN}) for simplicity. Each column has a \textit{NaN} value, and we replace it with the mean of the respective column values. The values in the dataset are normalized using the smallest and largest values of the samples and they all lie in the range 0 and 1 to ensure consistency in predictions. We resampled the dataset daily, incorporating the average of the per-minute values thereby generating a dataset with 1400 samples to predict \textit{global active power}.
From \figurename~\ref{fig:periodicity}, which shows the distribution of global\_active\_power per day, per week, per month, and per quarter, it is evident that the periodicity of the distribution decreases as the time interval increases.

\subsection{Backblaze Hard Disk Drive Dataset}\label{sec:hdd-dataset}
Before the surge in big data, model-driven approaches to Prognostic Health Management (PHM) depended on reactive maintenance and preventative maintenance techniques. In reactive maintenance, the hard disks are replaced only after failure resulting in disruption of normal operations until the drive is fixed. In preventative maintenance, the hard disks are replaced well before failure leading to the replacement of a fully functioning disk and resulting in wastage of resources. In data-driven approaches, predictive maintenance is performed using the Remaining Useful Life (RUL) metric. RUL is an important metric used to indicate the time to failure of any equipment. To prove the transferability of the attacks and defenses on any real-world dataset, we employ the data store from \citeA{Backblaze2023} housing the hard drive sensor readings for a period of 9 years from 2014 to 2022. We concentrate on the Seagate family of hard disk drives since they have the most amount of reliable data \cite{hdd-rohan}.

\begin{figure}[!htb]
  \centering
  \includegraphics[width=0.7\linewidth]{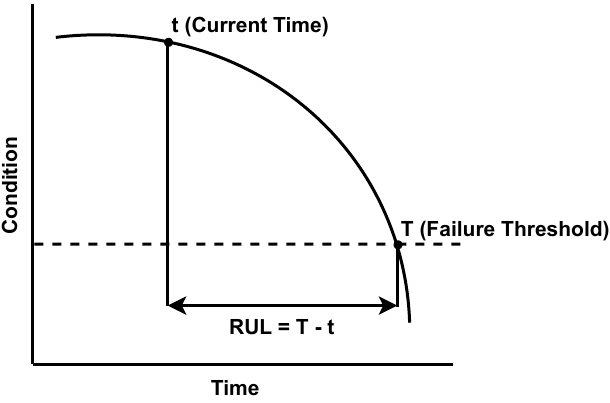}
  \caption{Remaining Useful Life curve.}
  \label{fig:rul}
\end{figure}

The dataset consists of attributes such as \textit{date, serial\_number, model, capacity, failure} and multiple \textit{S.M.A.R.T features}. S.M.A.R.T records various attributes of the hard disk drive. The date represents the time the reading is recorded in YYYY-MM-DD format, serial\_number, and model columns represent the serial number and model number assigned by the manufacturer. The model number of the dataset used is ST4000DM000 where ST stands for Seagate. The capacity column refers to the capacity of the HDD, and the final failure column consists of a binary 0/1 value which represents whether the hard disk drive has failed or not. Using the failure column, the RUL column is created to generate 5 different types of datasets each giving the time to failure of the disk starting from 5, 15, 25, 35, or 45 days. Remaining Useful Life can be calculated from the degradation curve shown in \figurename~\ref{fig:rul}.

If \textit{t} is the current time at which the disk is healthy and functioning properly, and \textit{T} is the time at which the disk fails (given by the failure column), then the RUL of the disk is given by \textit{T-t}. In other words, the day before the failure is marked as 1, the penultimate day before the failure is marked as 2, and so on up to 5, 15, 25, 35, and 45 days generating 5 different kinds of datasets with an increasing number of samples with an increase in look back days for the experiments. RUL is given by:
\begin{equation}
\text{RUL} = \text{Date of failure of HDD} - \text{First log date of HDD}
\end{equation}

The dataset can now be modeled as a multivariate time-series problem \cite{mohapatra2023tfbest}. The dataset is preprocessed by dropping any null or missing values and by normalizing the data values between 0 and 255 which is the standard proposed by Backblaze to account for the wide fluctuation of data values between 0 and $10^{14}$. 

\section{Training Processes}\label{sec:model_training}
We use a vanilla LSTM model \cite{lstm} on the Electricity dataset and an Encoder-Decoder LSTM model \cite{encoder-decoder} on the HDD dataset. The next two subsections consist of details about the experiments performed.

\subsection{Individual Household Power Consumption Dataset}\label{sec:uci_training}
We propose a vanilla LSTM model due to the fixed number of attributes and target values and because it is also the most prevalent model for multivariate time-series regression tasks on the Electricity dataset \cite{lstmelec1} \cite{lstmelec2} \cite{lstmelec3}.

The dataset is divided into 80\%, 10\%, and 10\% for the training, validation, and test sets respectively. We trained the vanilla LSTM model consisting of a sequential layer followed by 100 hidden nodes with the ReLu activation function. ReLU overcomes the problem of vanishing gradients by outputting a value equal to the input if the input is positive. 

We also added a 10\% dropout to regularize the network and prevent overfitting. Finally, we added a dense fully connected layer to the LSTM. Adam \cite{adam} is used as the optimizer and the metric defined is Root Mean Squared Error (RMSE). 

We divided the experiment into 3 parts to validate our findings and the behavior of the model:

\textit{1) Without using cross-validation:} We used the vanilla LSTM model to predict the \textit{global active power} target variable by using the same validation set to inform training.

\begin{figure}[!htb]
  \centering
  \includegraphics[width=0.7\linewidth]{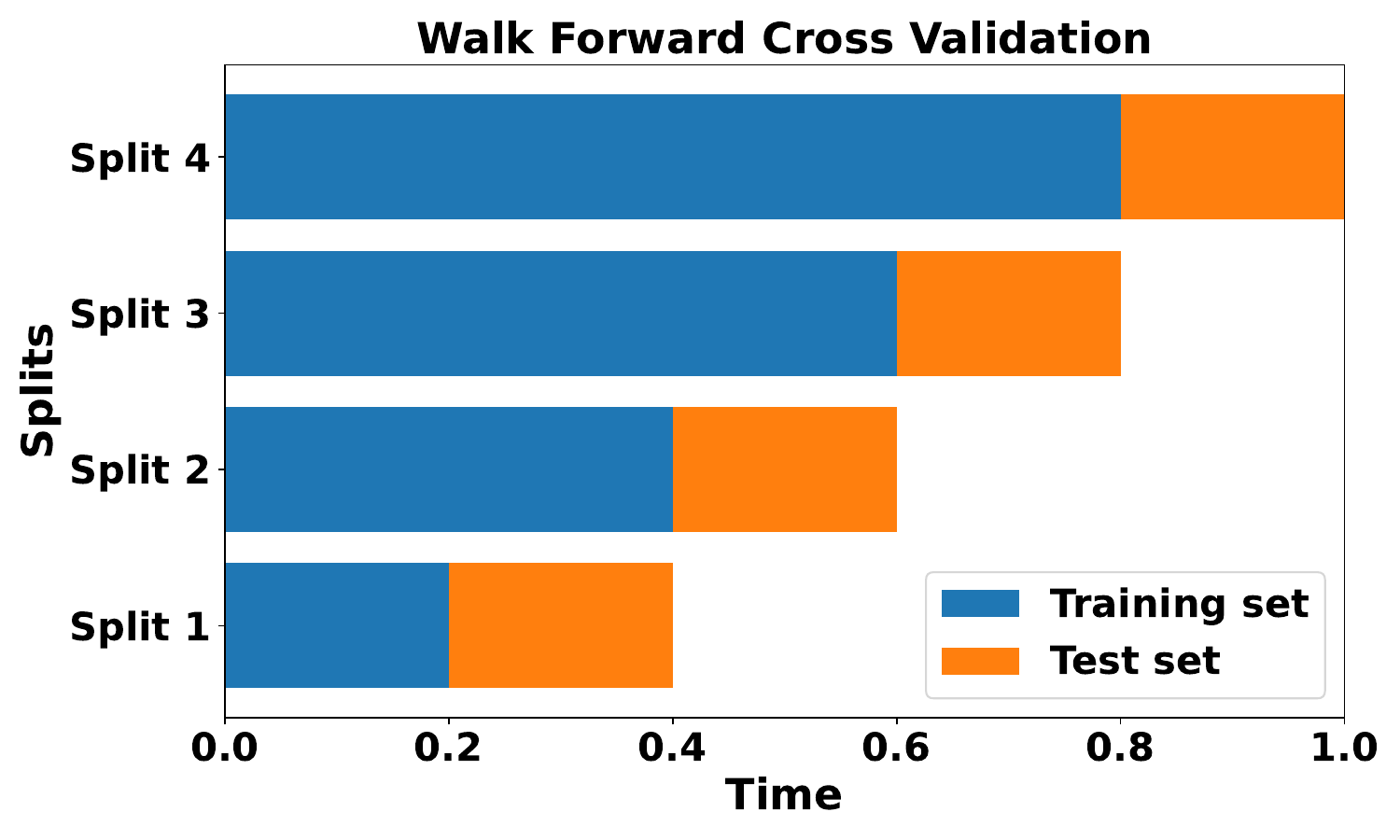}
  \caption{Walk-forward cross-validation.}
  \label{fig:walk_forward_validation}
\end{figure}

\textit{2) Using walk-forward cross-validation:} We repeated the experiment with 3, 5, and 10-fold cross-validation. Since classical k-fold and stratified cross-validation schemes shuffle the data and disrupt the order and seasonality of time-series input, we used walk-forward cross-validation. In walk-forward cross-validation, the first few data points in a finite window correspond to the training set and the next few data points correspond to the validation set. In the next iteration, more data points that were formerly in the validation set are included in the train set, and the window is expanded to include subsequent data points in the validation set as shown in \figurename~\ref{fig:walk_forward_validation}. 
Based on the amount of folds, this process is repeated and the average RMSE score is reported as the final training and validation RMSE scores. These k-fold cross validation schemes were also carried out by only looking back 1 day into the past to predict the next day's consumption. 

\begin{figure}[!htb]
  \centering
  \includegraphics[width=0.7\linewidth]{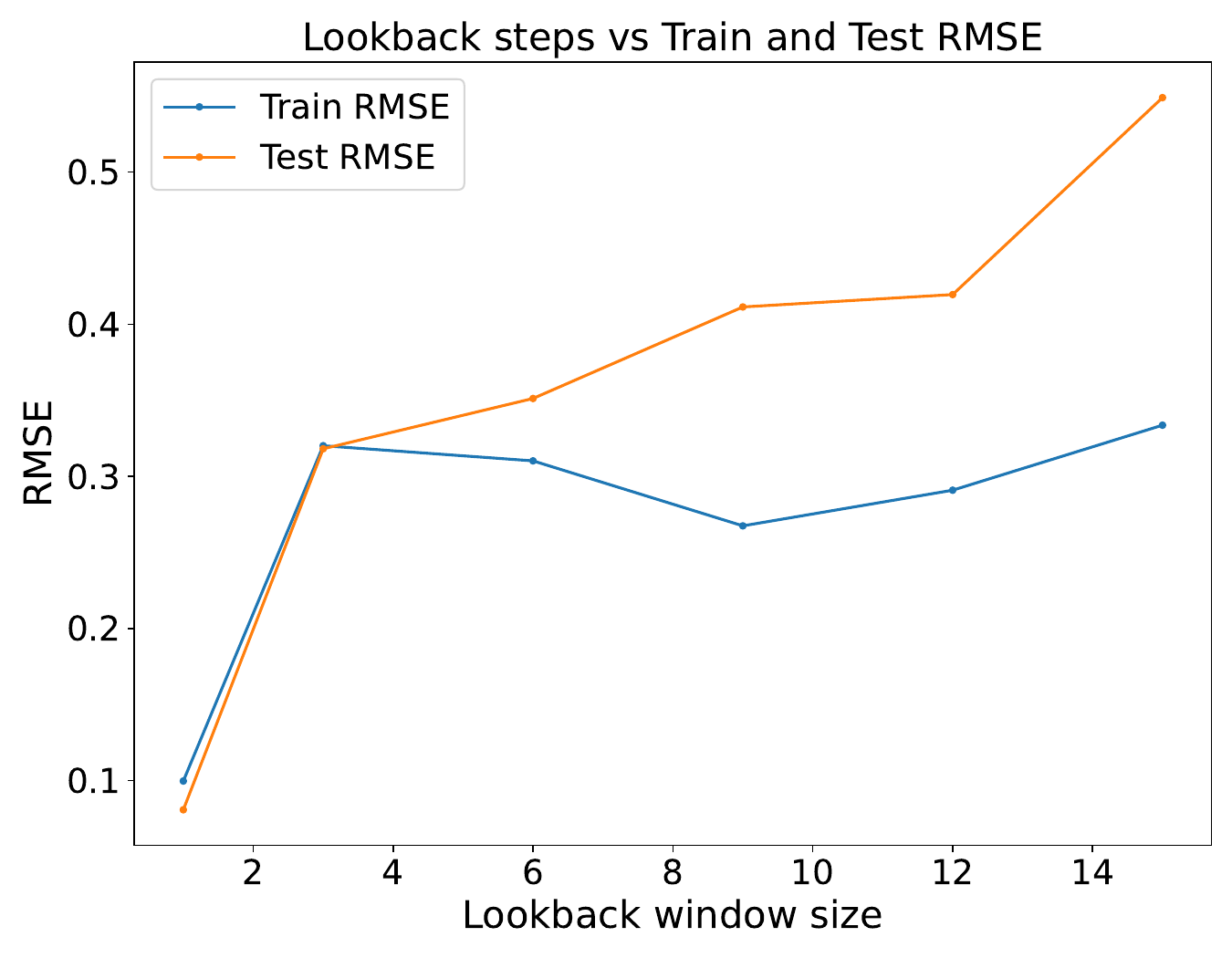}
  \caption{Error rate for varying look back window sizes.}
  \label{fig:lookback-graph}
\end{figure}

\textit{3) Using look-back:} The vanilla LSTM model trained so far looked back one day in the past to predict the future. We conducted experiments by increasing the look-back window size to allow the model to consider more samples from the past while making future predictions. The look-back window sizes we used in our experiments are 3, 6, 9, 12 and 15 days. The graph of the RMSE values for the train and test sets for varying look-back window sizes are shown in \figurename~\ref{fig:lookback-graph}. It is evident that as the look-back interval increases, the test set error increases, showing that looking further into the past worsens predictions. This can be attributed partly to the fact that LSTMs do not work well with long sequences and partly to the property inherent in the dataset, where increasing the time interval decreases the periodicity as shown in Fig.~\ref{fig:periodicity}. Since we have already down-sampled the dataset from minutes to days, samples further in the past do not contribute to the seasonal trends.

The RMSE scores of the train, validation, and test sets produced while running the experiments are tabulated in \tablename~\ref{tbl:electricity-results}. It is clear that the train RMSE is consistently higher than the test RMSE in all types of experiments performed. To eliminate sampling bias as a reason for this, we split the dataset into 60-20-20 to check if the test RMSE is still lower than the training RMSE. We found the test RMSE (0.0746) to still be lower than the train RMSE (0.1053) elucidating that the test set distribution mirrors the training set distribution due to the conspicuous seasonality inherent in the dataset making the process of predicting the \textit{global active power} easier, once the model has learned using the training set.
\begin{table}[!htb] \small
\caption{Experimental Results of the Vanilla LSTM Network on the Unpoisoned Dataset}
\label{tbl:electricity-results}
\centering
\begin{tabular}{l|l|l|l}
\hline \hline
\textbf{Experiments} & \textbf{Train} & \textbf{Validation} & \textbf{Test} \\ 
\hline \hline
Without using CV & 0.0997 & 0.0703 & 0.0807 \\  \hline
\textbf{Using 3-fold CV} & 0.1120 & 0.0849 & \textbf{0.0764}  \\ \hline
Using 5-fold CV & 0.1166 & 0.0858 & 0.0805  \\  \hline
Using 10-fold CV & 0.1156 & 0.0857 & 0.0805  \\ \hline
Using 3-day look-back & 0.3241 & 0.3073 & 0.3181  \\  \hline
Using 6-day look-back & 0.3102 & 0.2758 & 0.3512  \\  \hline
Using 9-day look-back & 0.2674 & 0.2744 & 0.4113  \\  \hline
Using 12-day look-back & 0.2909 & 0.1805 & 0.4195  \\ \hline 
Using 15-day look-back & 0.3336 & 0.1830 & 0.5490  \\ \hline
\end{tabular}
\end{table}

\begin{figure}[!htb]
  \centering
  \includegraphics[width=\linewidth]{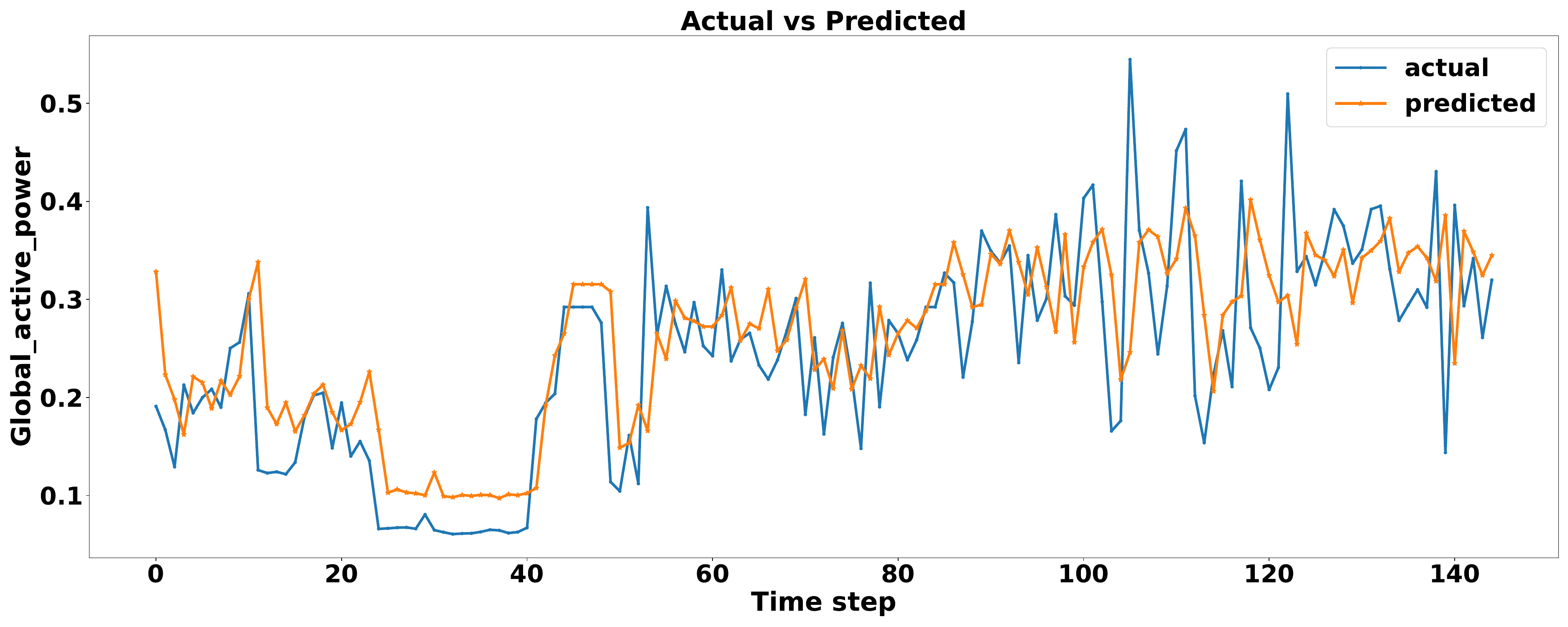}
  \caption{True vs. predicted values of the best LSTM model.}
  \label{fig:3-fold-graph}
\end{figure}

It is seen that 3-fold walk-forward cross-validation gives the best test RMSE result. \figurename~\ref{fig:3-fold-graph} shows the actual and predicted values. 

We conclude our experiments and select the 3-fold walk-forward cross-validation model without any feature selection as the model to conduct further experiments on. 

\subsection{Backblaze Hard Disk Drive Dataset}\label{sec:backblaze_training}
We propose an Encoder-Decoder LSTM \cite{encoder-decoder} as the model to learn the underlying distribution of the HDD dataset from Backblaze. We chose this model architecture because of the varying nature of the input sequence due to the addition of S.M.A.R.T features over the years, and also since we are aiming to predict a sequence of RUL values given the sequence of inputs. This Encoder-Decoder model generates the most likely sequence given a sequence of data. The encoder scans the input and outputs a constant-size array called the context vector, and the decoder reads from the context vector.

Similar to the electricity dataset, this dataset is split following the 80-10-10 rule for the train, validation, and test sets respectively. Our Encoder-Decoder LSTM model consists of a Sequential Layer of one hundred hidden units with ReLu activation. We look 5, 15, 25, 35, and 45 days into the past to check the effect of feeding more past data to the model. Since we are looking back ‘t’ periods, the output of the encoder is repeated ‘t’ times before passing it through another ReLu layer with 100 units
We added a 10\% dropout to prevent overfitting and help in generalization. A dense layer is added to every period of the decoder's output series using the \textit{Time Distributed} wrapper thereby enabling the model to predict the RUL for each time step. Adam optimizer is utilized to fine-tune the weights during training, and RMSE is used as metrics. For higher lookback time steps, the gradients explode and to prevent NaN metrics, the gradients are clipped if they exceed 0.5. 

\begin{figure}[!ht]
  \centering
  \includegraphics[width=0.7\linewidth]{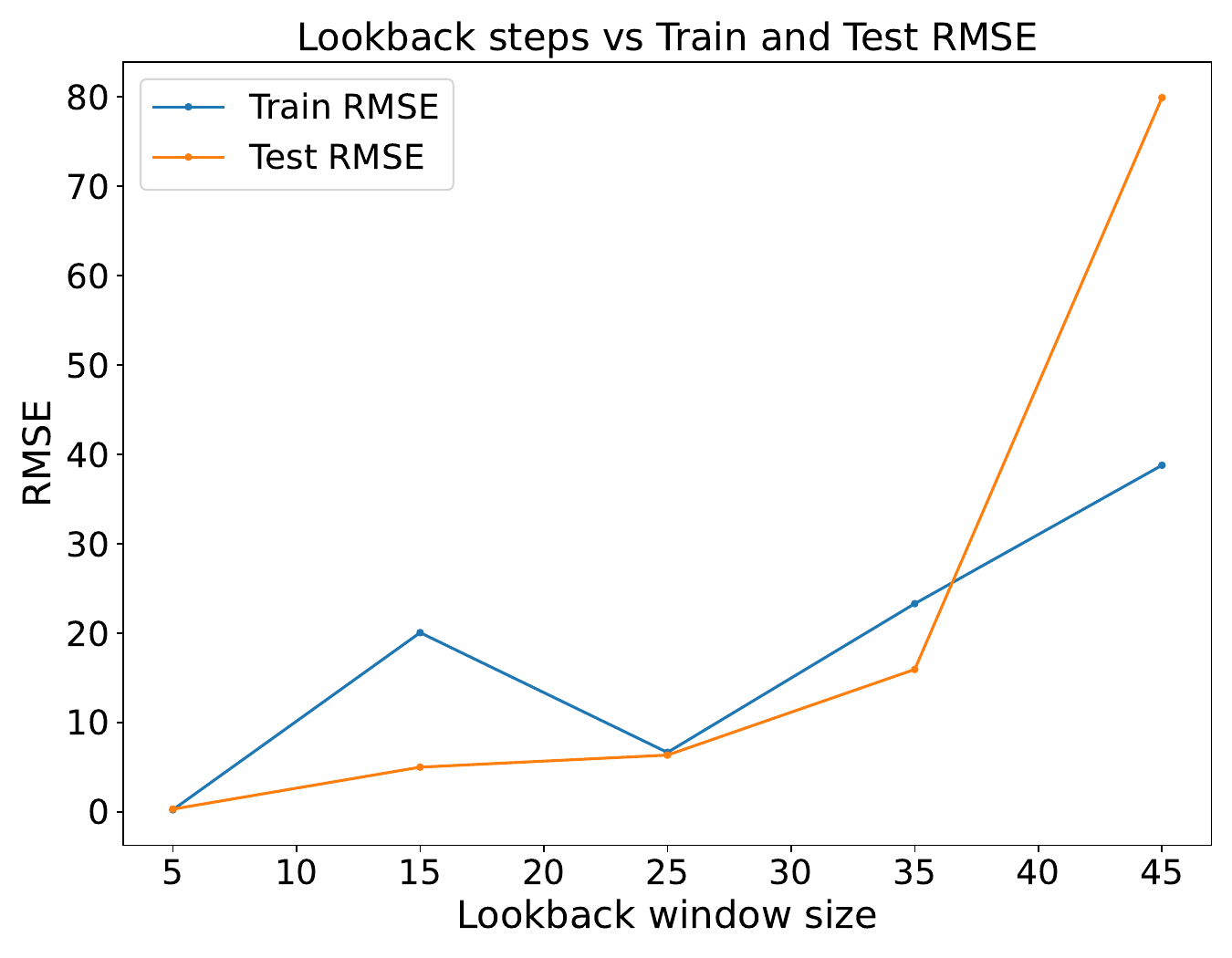}
  \caption{Error rate for varying look back window sizes.}
  \label{fig:hdd-lookback-graph}
\end{figure}

We divide our experiments into two to choose the top-performing model to execute the attacks and defenses as follows:

\textit{1) Using look-back:} We facilitate the learning of the Encoder-Decoder LSTM model by providing the five different kinds of datasets generated in Section~\ref{sec:hdd-dataset}. From \figurename~\ref{fig:hdd-lookback-graph}, it is clear that increasing the look-back window exacerbates predictions since the main disadvantage of an Encoder-Decoder LSTM is its inability to handle long sequences. We conduct cross-validation experiments on the 5-day look-back dataset.

\textit{2) Using walk-forward cross validation}: Like the electricity dataset, we perform 3, 5, and 10-fold walk-forward cross-validation on the 5-day look-back dataset. The results are tabulated in \tablename~\ref{tbl:hdd-results}.
\begin{table}[!htb] \small
\caption{Results of the LSTM Network on the Unpoisoned Dataset}
\label{tbl:hdd-results}
\centering
\begin{tabular}{l|l|l|l}
\hline \hline
\textbf{Experiments} & \textbf{Train} & \textbf{Validation} & \textbf{Test} \\ 
\hline \hline
Using 3-fold CV & 0.0677 & 1.1446 & 0.3084  \\ \hline
Using 5-fold CV & 0.0808 & 0.4718 & 0.2085  \\  \hline
\textbf{Using 10-fold CV} & 0.1065 & 0.8493 & \textbf{0.0715}  \\ \hline
Using 5-day look-back & 0.2156 & 1.0192 & 0.2886  \\  \hline
Using 15-day look-back & 20.0390 & 6.0372 & 4.9856 \\  \hline
Using 25-day look-back & 6.6315 & 7.5154 & 6.3345  \\  \hline
Using 35-day look-back & 23.2777 & 19.0839 & 15.9206  \\  \hline
Using 45-day look-back & 38.7680 & 52.4585 & 79.9001  \\ \hline
\end{tabular}
\end{table}

\begin{figure}[!htb]
  \centering
  \includegraphics[width=\linewidth]{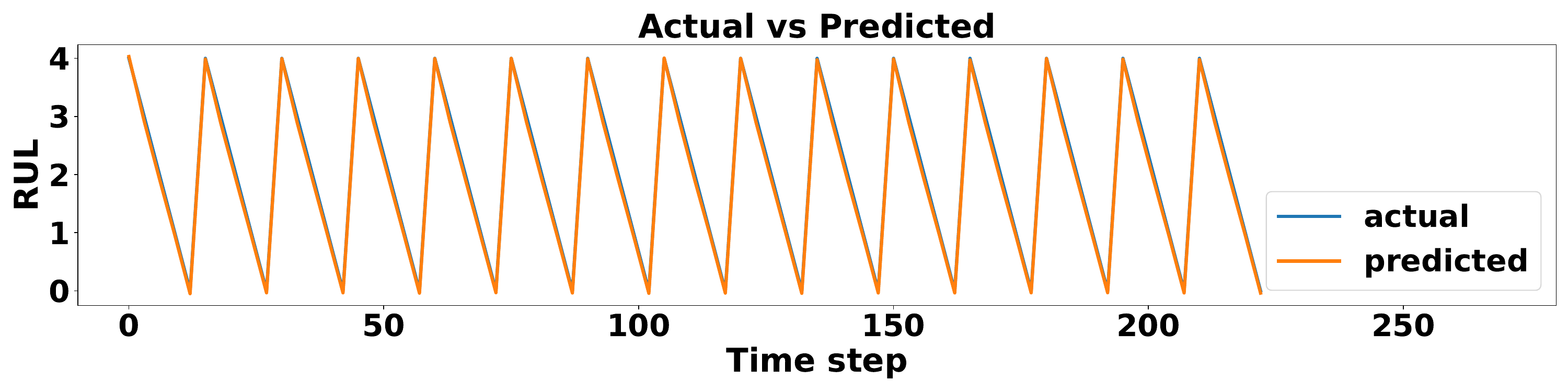}
  \caption{True vs. predicted values of the best Encoder-Decoder LSTM.}
  \label{fig:10-fold-graph}
\end{figure}

It is seen that 10-fold walk-forward cross-validation on the dataset which looks back 5 days gives the best results. \figurename~\ref{fig:10-fold-graph} shows the true versus predicted values of this Encoder-Decoder LSTM model.

We choose the 10-fold cross validated 5-day look-back model as the final one to attack and defend against. 

\section{Attacks and Defense Strategies}\label{sec:overview}
We illustrate the process of performing adversarial attacks and adversarial defenses in this section.
\subsection{Overview of Process Flow}
\begin{figure}[!htb]
  \centering
  \includegraphics[width=\linewidth]{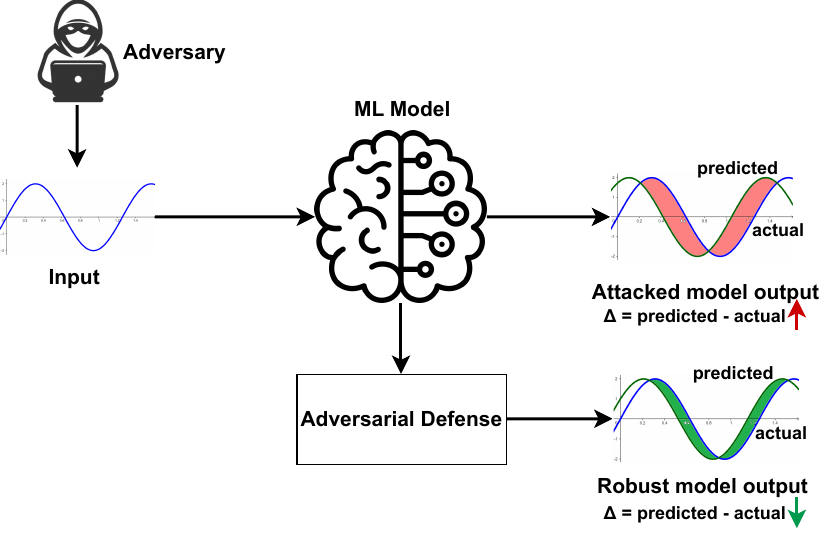}
  \caption{Overall System Architecture.}
  \label{fig:process-flow}
\end{figure}

We demonstrate that the adversarial attacks are successful and exploit the sequential nature of deep learning networks and their susceptibility to adversarial perturbations using FGSM, and BIM attacks. Once the best LSTM models selected in Section~\ref{sec:uci_training} and Section~\ref{sec:backblaze_training} succumb to the two types of attacks mentioned above, we perform adversarial defenses using data augmentation at the data plane, and layer-wise perturbations at the gradient plane to inform the training process. This ensures model robustness to adversarial attacks. The entire process flow is summarized in \figurename~\ref{fig:process-flow}.

\subsection{Adversarial Attacks}\label{sec:adversarial-attacks}
We perform two types of adversarial perturbations of the data.

\subsubsection{Fast Gradient Sign Method (FGSM)}
\cite{fgsm} proposed FGSM to perform adversarial perturbations on CNNs for image data. The FGSM algorithm adds disturbances to the input in the direction of the gradients with regard to the loss function of the data. We have extended the algorithm to multivariate time-series datasets described earlier. The equation for FGSM attack is:
\begin{equation}
X_{\text{adv}} = \text{X} + \epsilon \cdot \text{sign}\left(\nabla J(f, \text{X}, y)\right)
\end{equation}
where $X_{\text{adv}}$ is the disturbed input, $X$ is the actual input, $\epsilon$ is a constant (perturbation intensity), $\text{sign}(\cdot)$ computes the sign of the gradient, $J(f, X, y)$ is the gradient of $J$ with regard to $X$, $f$ is the neural network, and $y$ is the actual output.

\subsubsection{Basic Iterative Method (BIM)}
\cite{bim} proposed BIM which applies the FGSM attack multiple times. Since in each iteration, the attack forces the model to add noise or perturbation in the direction of the gradients with regard to the loss function, this attack mechanism is generally considered to be more powerful than FGSM. We have used BIM to attack the best-performing vanilla LSTM and Encoder-Decoder LSTM models selected earlier. Its equation is given as follows:
\begin{equation}
X_{\text{adv}} = X + \alpha \cdot \text{sign}(\nabla J(f, X, y))
\end{equation}
\begin{equation}
X_{\text{adv}} = \min(X + \epsilon, \max(X - \epsilon, X_{\text{adv}}))
\label{eq:bim}
\end{equation}
where $X_{\text{adv}}$ is the disturbed input, $X$ is the actual input, $\alpha$ is the step size, $\epsilon$ is a constant (perturbation intensity), $\text{sign}(\cdot)$ computes the sign of the gradient, $J(f, X, y)$ is the gradient of $J$ with regard to $X$, $f$ is the neural network, and $y$ is the actual output.

\subsection{Adversarial Defenses}
We enumerate the two types of adversarial defenses performed, in this section.

\subsubsection{Data Augmentation-based Adversarial Training (DAAT)}
\begin{figure}[!htb]
  \centering
  \includegraphics[width=\linewidth]{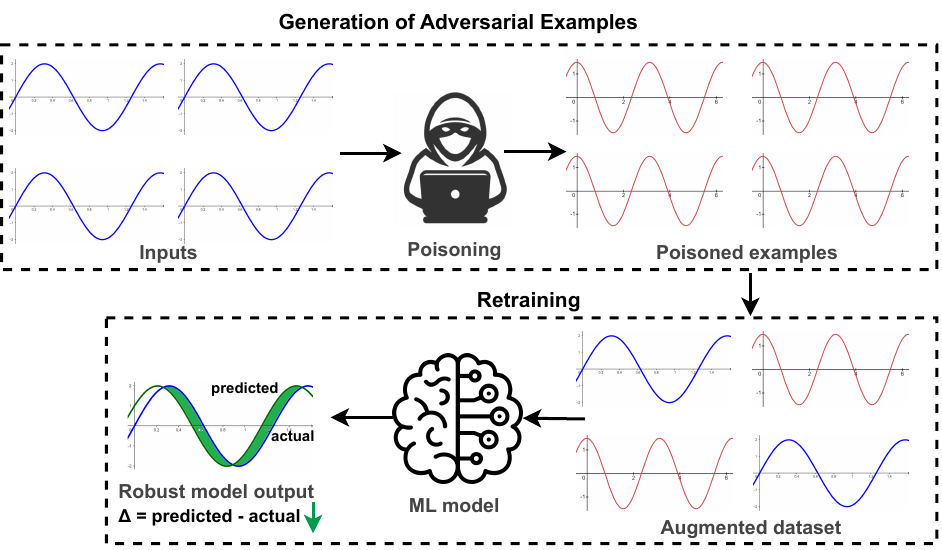}
  \caption{System Architecture of Data Augmentation-based Adversarial Training (DAAT).}
  \label{fig:daat}
\end{figure}

DAAT is a naïve process of using adversarial attacks to create adversarial examples and augmenting the dataset to incorporate these examples during the training of the deep learning models \cite{fgsm}. This mechanism is used after anticipating the types of perturbations the adversary can use during the attack. DAAT provides the deep learning models with prior information necessary to stay resilient to attacks during the training process. We used the adversarial attacks introduced earlier to augment the dataset with adversarial examples for different values of perturbation magnitude epsilon. Then we trained a robust classifier on the augmented dataset. This robust classifier is more resistant to adversarial attacks since it has been trained to predict the right values, given the perturbed values in its training set. The system architecture of DAAT is shown in \figurename~\ref{fig:daat}.

\subsubsection{Layer-wise Perturbation-Based Adversarial Training (LPAT)}
\begin{figure}[!htb]
  \centering
  \includegraphics[width=\linewidth]{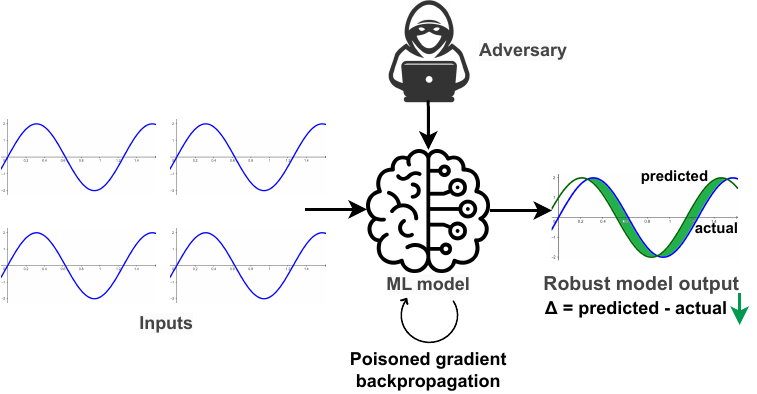}
  \caption{System Architecture of Layer-wise Perturbation-based Adversarial Training (LPAT).}
  \label{fig:lpat}
\end{figure}

\cite{lpat} proposed LPAT for hard drive health prediction, where they train an LSTM network to handle class imbalances better. This robust LSTM network is trained by going through two rounds of feed-forward and backpropagation in each iteration. The first round is similar to any neural network architecture where the outputs are computed in the forward propagation step, and the gradients are updated in the backpropagation step. In the second round, however, the gradients in each layer are perturbed using FGSM and BIM attacks before the forward and backward propagation steps are performed again. We extend this to include a deterministic gradient update, where a predetermined value is used always as the magnitude of gradient perturbation, and a stochastic gradient update, where any random value between a specified range is used to perturb the gradients. The system architecture of LPAT is shown in \figurename~\ref{fig:lpat}.

\section{Results}\label{sec:results}
In this section, we disclose the findings from carrying out the attacks and defenses on the final deep learning models selected in Section~\ref{sec:exp-setup}.
\subsection{Individual Household Power Consumption Dataset}\label{sec:elec-results}
\subsubsection{Results of Adversarial Attacks}\label{sec:elec-attack-results}
We contaminate the inputs to the vanilla LSTM model during the evaluation stage using the FGSM and BIM attacks mentioned in Section~\ref{sec:adversarial-attacks}. For this experiment with the electricity dataset, we find that smaller orders of perturbation magnitude between 0.05 and 0.25, incremented in steps of 0.05 give test set RMSE values in the range of 0.117 and 0.3746 for FGSM, and between 0.1287 and 0.4575 for BIM.  

$\alpha$ parameter in Equation~\ref{eq:bim} is always set to 0.01 for all the experiments conducted. The total iterations I is given as follows:
\begin{equation}
I = \min\left(4 + \frac{\epsilon}{\alpha}, 1.25 \cdot \frac{\epsilon}{\alpha}\right)
\label{eq:bim-iterator}
\end{equation}
where $\epsilon$ is the degree of fluctuation and $\alpha$ is the step size.

The results of the attacks on the electricity dataset are tabulated in Table~\ref{tbl:elec-attack}. It is seen that RMSE increases with increasing $\epsilon$. 
\begin{table}[!htb] \small
\caption{RMSE Values on the Test Electricity Dataset after Attack}
\label{tbl:elec-attack}
\centering
\begin{tabular}{l|l|l|l}
\hline \hline
\multicolumn{1}{l|}{\textbf{Attack Type}} & \textbf{$\epsilon$} & \textbf{Attack RMSE} & \textbf{\% increase in error} \\ \hline \hline
\multirow{4}{*}{FGSM} & \multicolumn{1}{c|}{0.05} & 0.117 & 54.14 \\  \cline{2-4}
                      & \multicolumn{1}{c|}{0.1} & 0.1741 & 127.88 \\  \cline{2-4}
                      & \multicolumn{1}{c|}{0.15} & 0.237 & 210.21 \\  \cline{2-4}
                      & \multicolumn{1}{c|}{0.2} & 0.3039 & 297.79 \\ \cline{2-4}
                      & \multicolumn{1}{c|}{0.25} & 0.3746 & 390.31 \\ \hline   
\multirow{4}{*}{BIM} & \multicolumn{1}{c|}{0.05} & 0.1287 & 68.46 \\  \cline{2-4}
                      & \multicolumn{1}{c|}{0.1} & 0.2029 & 165.58 \\  \cline{2-4}
                      & \multicolumn{1}{c|}{0.15} & 0.2866 & 275.13 \\  \cline{2-4}
                      & \multicolumn{1}{c|}{0.2} & 0.3776 & 394.24 \\ \cline{2-4}
                      & \multicolumn{1}{c|}{0.25} & 0.4575 & 498.82 \\ \hline
\end{tabular}
\end{table}

\begin{figure}[!htb]
  \centering
  \includegraphics[width=0.7\linewidth]{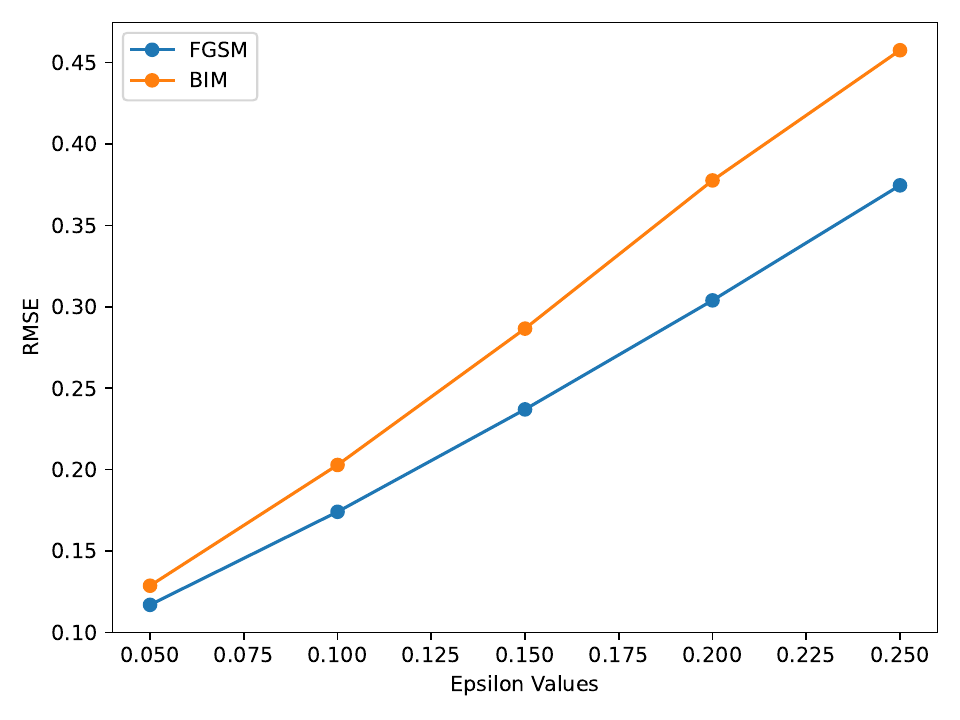}
  \caption{Comparison of test RMSE for FGSM and BIM with varying $\epsilon$ on the Electricity dataset.}
  \label{fig:elec-attack-rmse}
\end{figure}

We see that BIM is a stronger attack than FGSM from \figurename~\ref{fig:elec-attack-rmse}.

\begin{figure}[!htb]
  \centering
  \includegraphics[width=\linewidth]{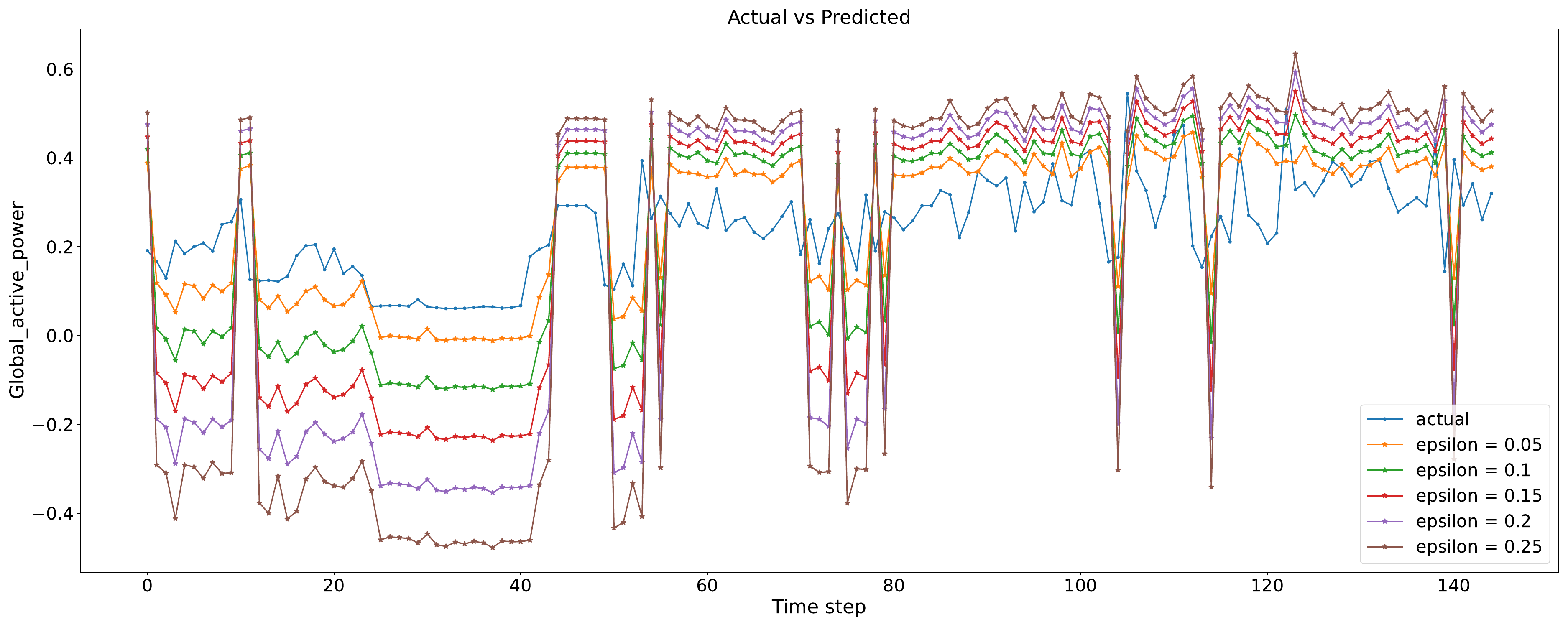}
  \caption{Predictions on the FGSM perturbed electricity test dataset for varying $\epsilon$.}
  \label{fig:elec-fgsm-curve}
\end{figure}

\begin{figure}[!htb]
  \centering
  \includegraphics[width=\linewidth]{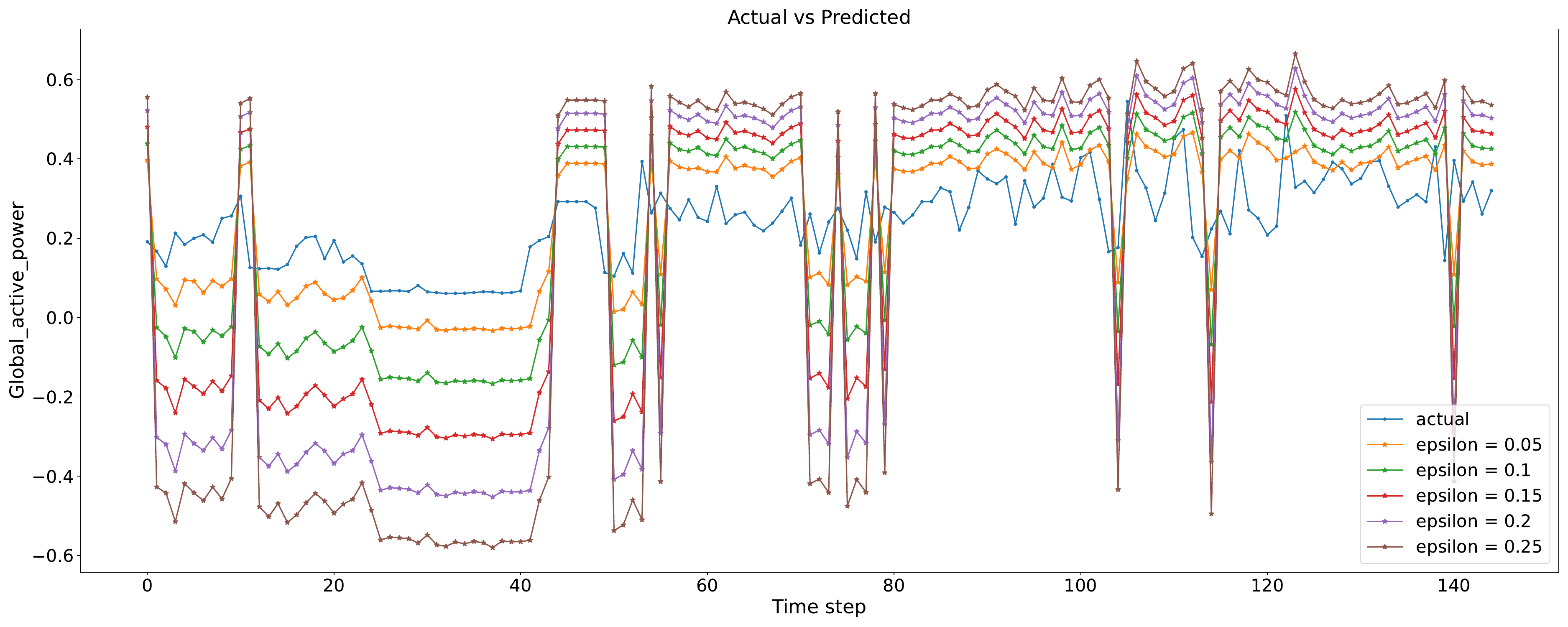}
  \caption{Predictions on the BIM perturbed electricity test dataset for varying $\epsilon$.}
  \label{fig:elec-bim-curve}
\end{figure}

\begin{figure}[!htb]
  \centering
  \includegraphics[width=\linewidth]{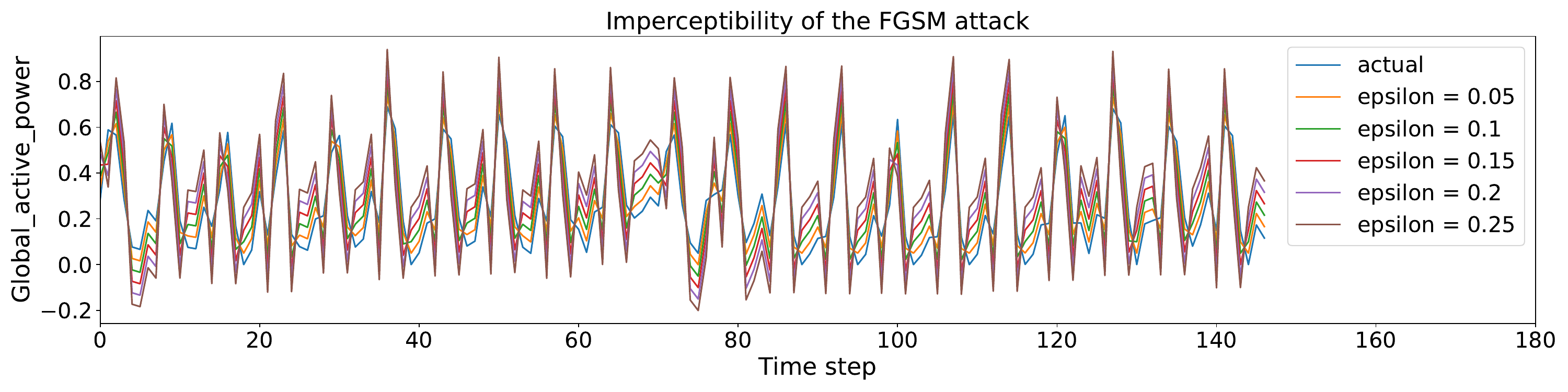}
  \caption{Imperceptibility of the FGSM attack on the electricity dataset.}
  \label{fig:elec-fgsm-imperceptible}
\end{figure}

\begin{figure}[!htb]
  \centering
  \includegraphics[width=\linewidth]{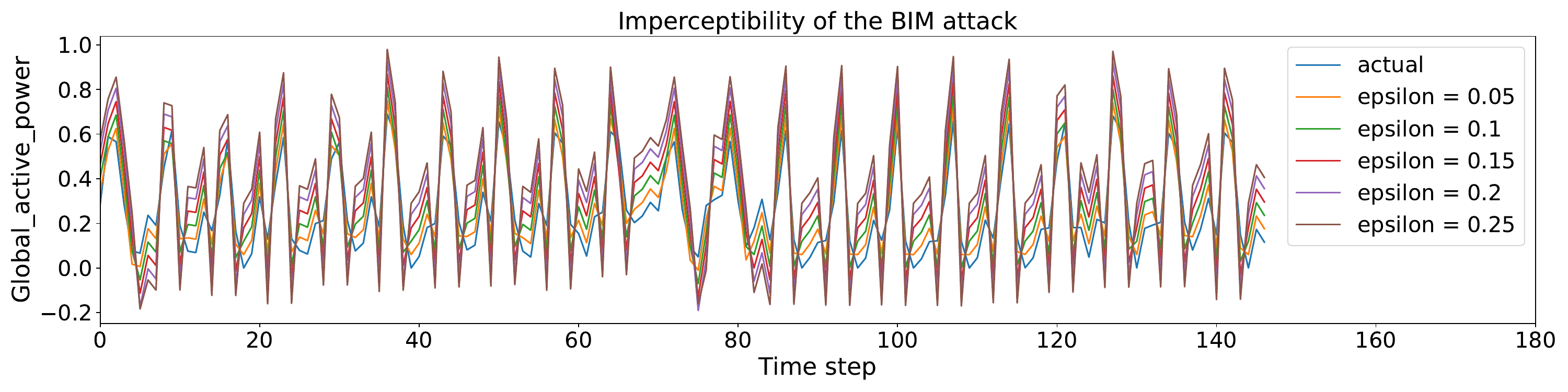}
  \caption{Imperceptibility of the BIM attack on the electricity dataset.}
  \label{fig:elec-bim-imperceptible}
\end{figure}

\figurename~\ref{fig:elec-fgsm-curve} and \figurename~\ref{fig:elec-bim-curve} show the estimations on the altered evaluation dataset after the FGSM and BIM attacks respectively. \figurename~\ref{fig:elec-fgsm-imperceptible} and \figurename~\ref{fig:elec-bim-imperceptible} show the changes in a subset of input data after the FGSM and BIM attacks respectively. 

It is obvious that the changes in the train set distribution for all $\epsilon$ values are almost imperceptible to the naked eye, but the predictions on the test set vary widely demonstrating the danger of these attacks.

\subsubsection{Results of Adversarial Defenses}\label{sec:elec-defense-results}
We carry out two different types of adversarial defense strategies to harden the best-performing vanilla LSTM model on the electricity dataset. In DAAT, we augment the dataset with perturbed input attributes created by performing adversarial perturbations on the original feature samples. We train the vanilla LSTM network to learn the fluctuations in the features and predict the correct target. The different RMSE values obtained after performing DAAT on the electricity dataset are tabulated in \tablename~\ref{tbl:elec-daat}. Although BIM is a stronger attack than FGSM, DAAT on the electricity dataset is slightly more resilient to BIM attack. The observed phenomenon can be ascribed to the broader range of variations in the test features perturbed by BIM enhancing the model's capability to identify noisy outliers within the underlying distribution more effectively.

\begin{table}[!htb] \small
\caption{Results after Performing DAAT on the Electricity Dataset}
\label{tbl:elec-daat}
\centering
\begin{tabular}{l|l|l}
\hline \hline
\textbf{Type of Attack} & \textbf{RMSE Metrics} & \textbf{Values} \\ \hline \hline
\multirow{4}{*}{FGSM} & Train & 0.0994 \\ \cline{2-3}
 & Validation & 0.1271 \\ \cline{2-3}
 & Test (clean data) & 0.0761 \\ \cline{2-3}
 & Test (poisoned data) & 0.0847 \\ \hline
\multirow{4}{*}{BIM} & Train & 0.0986 \\ \cline{2-3}
 & Validation & 0.1279 \\ \cline{2-3}
 & Test (clean data) & 0.0802 \\ \cline{2-3}
 & Test (poisoned data) & 0.0949 \\ \hline
\end{tabular}
\end{table}

In the second method, we harden the vanilla LSTM model by perturbing the gradients of the model during backpropagation, allowing it to account for the outliers better. We introduced two types of gradient update - deterministic and stochastic. In the deterministic update, the mean of $\epsilon$ (in this case 0.15) is used, whereas in the stochastic process, we choose from a range of $\epsilon$ (in this case between 0.05 and 0.25). The various RMSE metrics obtained after using LPAT on the electricity dataset are tabulated in \tablename~\ref{tbl:elec-lpat}. 

\begin{table}[!htb] \small
\caption{Results after Performing LPAT on the Electricity Dataset}
\label{tbl:elec-lpat}
\centering
\begin{tabular}{l|l|l|l}
\hline \hline
    \textbf{Attack Type} & \textbf{Training Type} & \textbf{RMSE Metrics} & \textbf{Value} \\ \hline \hline
    \multirow{4}{*}{FGSM} & \multirow{3}{*}{Deterministic} & Train & 0.1413 \\ \cline{3-4}
    & & Test & 0.0974 \\ \cline{2-4} 
    & \multirow{3}{*}{Stochastic} & Train & 0.1424 \\ \cline{3-4}
    & & Test & 0.0979 \\ \hline
    \multirow{4}{*}{BIM} & \multirow{3}{*}{Deterministic} & Train & 0.1304 \\ \cline{3-4}
    & & Test & 0.0985 \\ \cline{2-4} 
    & \multirow{3}{*}{Stochastic} & Train & 0.1389 \\ \cline{3-4}
    & & Test & 0.101 \\ \hline
\end{tabular}
\end{table}

\begin{figure}[!htb]
  \centering
  
  \includegraphics[width=0.85\linewidth]{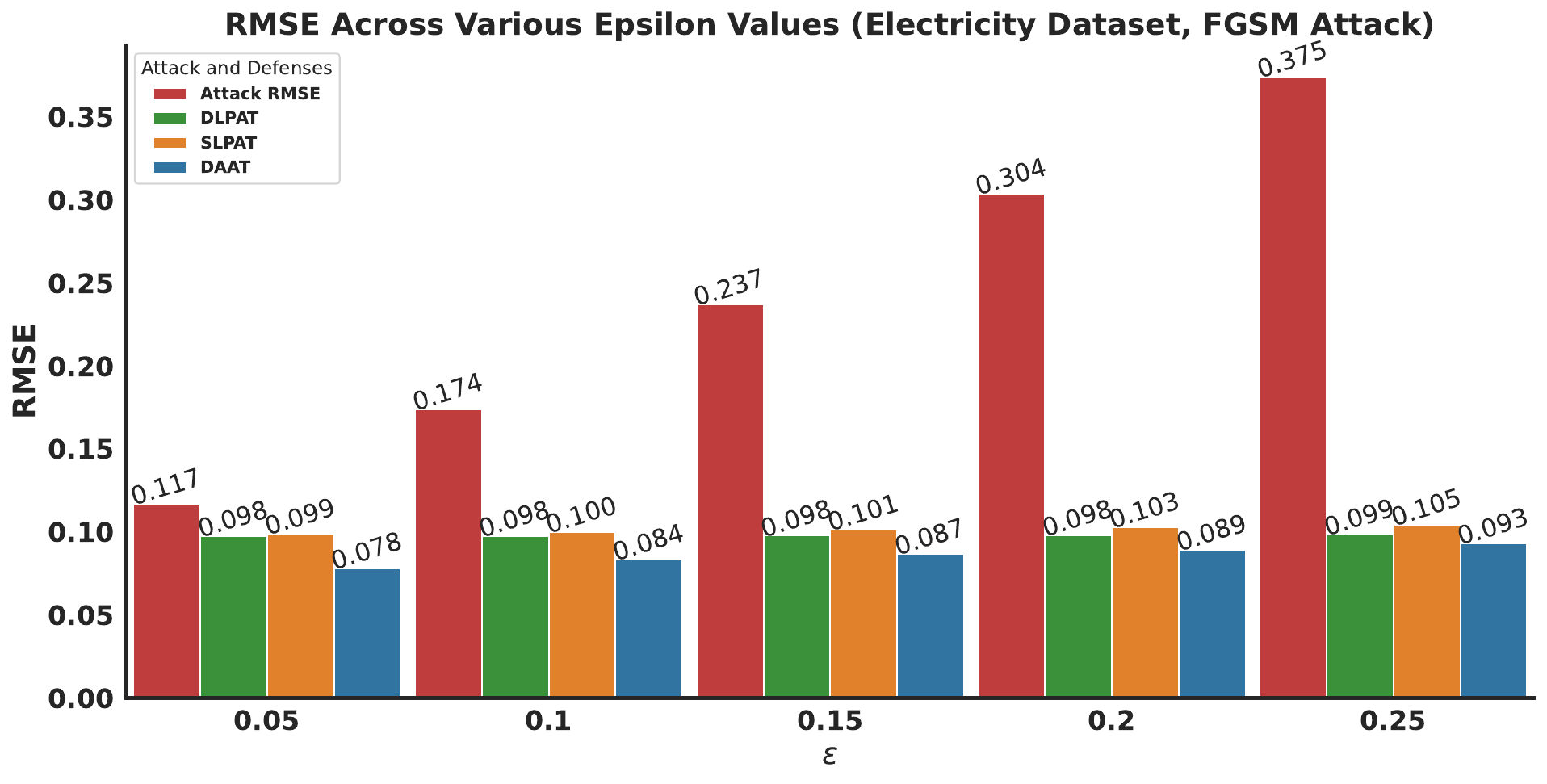}
  \caption{RMSE of FGSM attack and defenses for electricity test set perturbed by varying $\epsilon$.}
  \label{fig:elec-fgsm}
\end{figure}

\begin{figure}[!htb]
  \centering

  \includegraphics[width=0.85\linewidth]{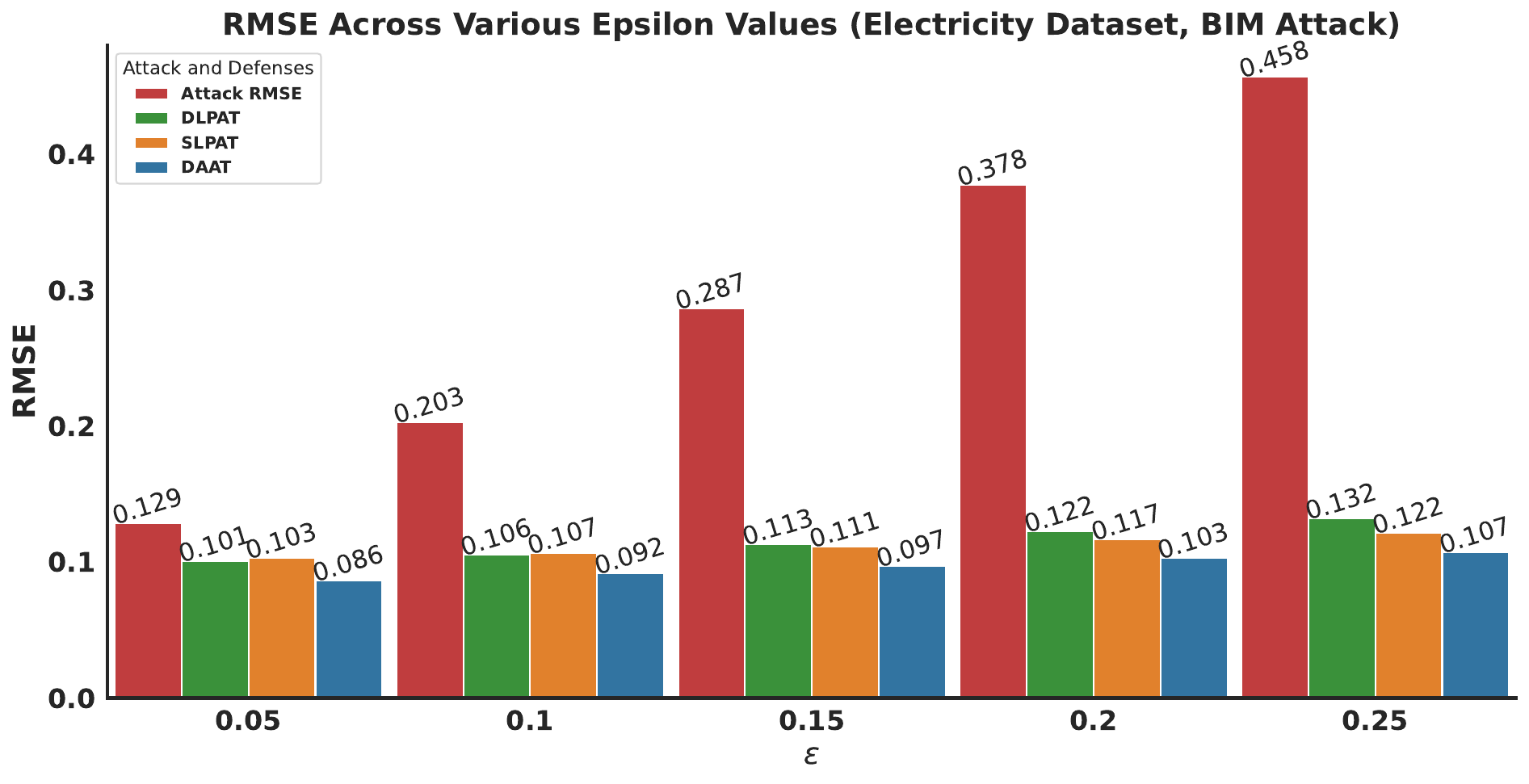}
  \caption{RMSE of BIM attack and defenses for electricity test set perturbed by varying $\epsilon$.}
  \label{fig:elec-bim}
\end{figure}

\begin{figure}[!htb]
  \centering
 
  \includegraphics[width=0.85\linewidth]{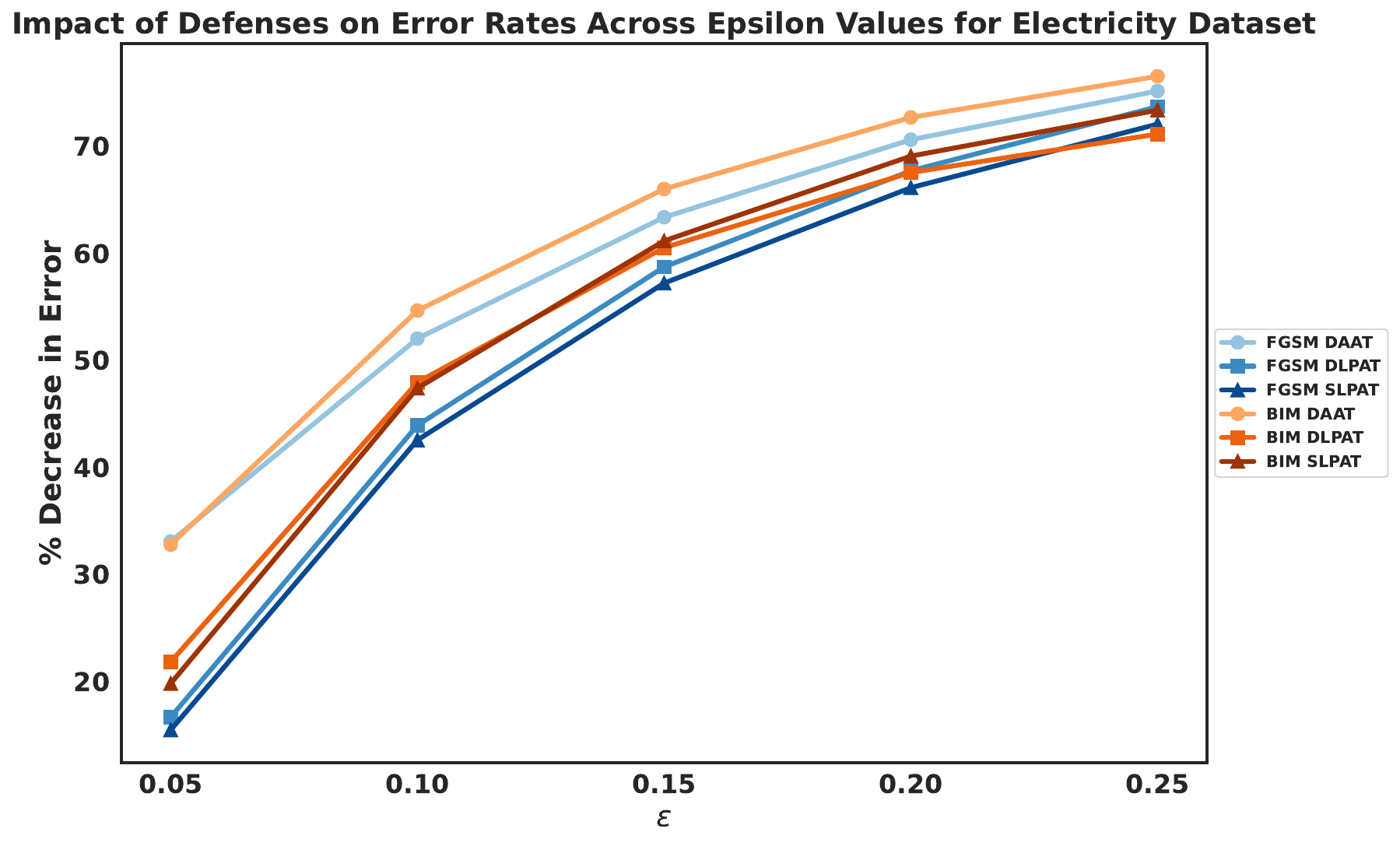}
  \caption{Percentage Reduction in Error Values after Adversarial Defense on the Electricity Dataset}
  \label{fig:percent-decrease-elec}
\end{figure}

The RMSE values after the FGSM and BIM attacks and defenses are shown in \figurename~\ref{fig:elec-fgsm} and \figurename~\ref{fig:elec-bim} respectively. \figurename~\ref{fig:percent-decrease-elec} a line chart representing the percentage decrease in error on the electricity dataset for different $\epsilon$ after performing the defenses showing that BIM DAAT followed by FGSM DAAT gives the maximum reduction in error during training. This is followed by BIM-based DLPAT and SLPAT techniques and finally FGSM-based DLPAT and SLPAT.


\subsection{Backblaze Hard Disk Drive Dataset}
\subsubsection{Results of Adversarial Attacks}\label{sec:hdd-attack-results}
We poison the test inputs to the best-performing model identified in Section~\ref{sec:backblaze_training} using the FGSM and BIM attacks. We used epsilon ranging from 3 to 11 incremented in steps of 2 such as 3, 5, 7, 9, and 11 since the Encoder-Decoder LSTM model is inherently robust to lower $\epsilon$ resulting in only a 0.92\% to 3.87\% increase in error rate for FGSM and a 1.12\% to 4.7\% increase in error rate for BIM for $\epsilon$ between 0.05 and 0.2.

$\alpha$ in Equation~\ref{eq:bim} is set to 0.01 for all the experiments. The number of iterations I is given by Equation~\ref{eq:bim-iterator}.

The results for the perturbations performed on the HDD dataset tabulated in Table~\ref{tbl:hdd-attack} show that RMSE increases with increasing $\epsilon$.
\begin{table}[!htb] \small
\caption{RMSE Values on the Test Hard Disk Drive (HDD) Dataset after Attack}
\label{tbl:hdd-attack}
\centering
\begin{tabular}{l|l|l|l}
\hline \hline
\multicolumn{1}{l|}{\textbf{Attack Type}} & \textbf{$\epsilon$} & \textbf{Attack RMSE} & \textbf{\% increase in error} \\ \hline \hline
\multirow{4}{*}{FGSM} & \multicolumn{1}{c|}{3} & 0.124 & 73.43 \\  \cline{2-4}
                      & \multicolumn{1}{c|}{5} & 0.1683 & 135.39 \\  \cline{2-4}
                      & \multicolumn{1}{c|}{7} & 0.2154 & 201.26 \\  \cline{2-4}
                      & \multicolumn{1}{c|}{9} & 0.2605 & 264.34 \\ \cline{2-4}
                      & \multicolumn{1}{c|}{11} & 0.298 & 316.78 \\ \hline  
\multirow{4}{*}{BIM} & \multicolumn{1}{c|}{3} & 0.1344 & 87.97 \\  \cline{2-4}
                      & \multicolumn{1}{c|}{5} & 0.1904 & 166.29 \\  \cline{2-4}
                      & \multicolumn{1}{c|}{7} & 0.2505 & 250.35 \\  \cline{2-4}
                      & \multicolumn{1}{c|}{9} & 0.3075 & 330.07 \\ \cline{2-4}
                      & \multicolumn{1}{c|}{11} & 0.3609 & 404.76 \\ \hline
\end{tabular}
\end{table}

\begin{figure}[!htb]
  \centering
  \includegraphics[width=0.7\linewidth]{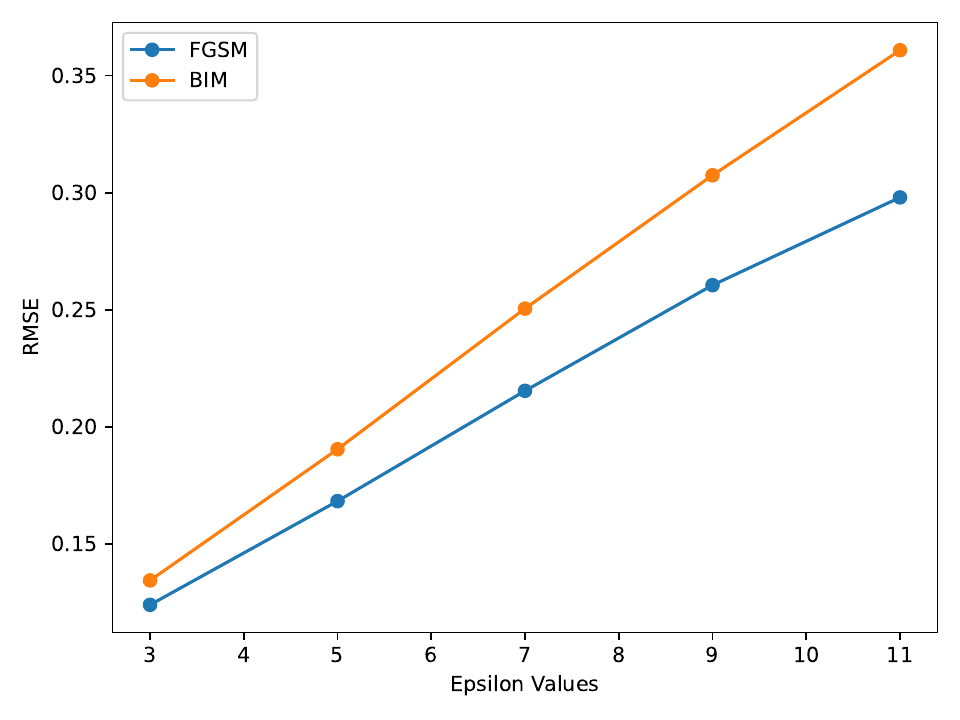}
  \caption{Comparison of test RMSE for FGSM and BIM with varying $\epsilon$ on the HDD dataset.}
  \label{fig:hdd-attack-rmse}
\end{figure}

\begin{figure}[!htb]
  \centering
  \includegraphics[width=\linewidth]{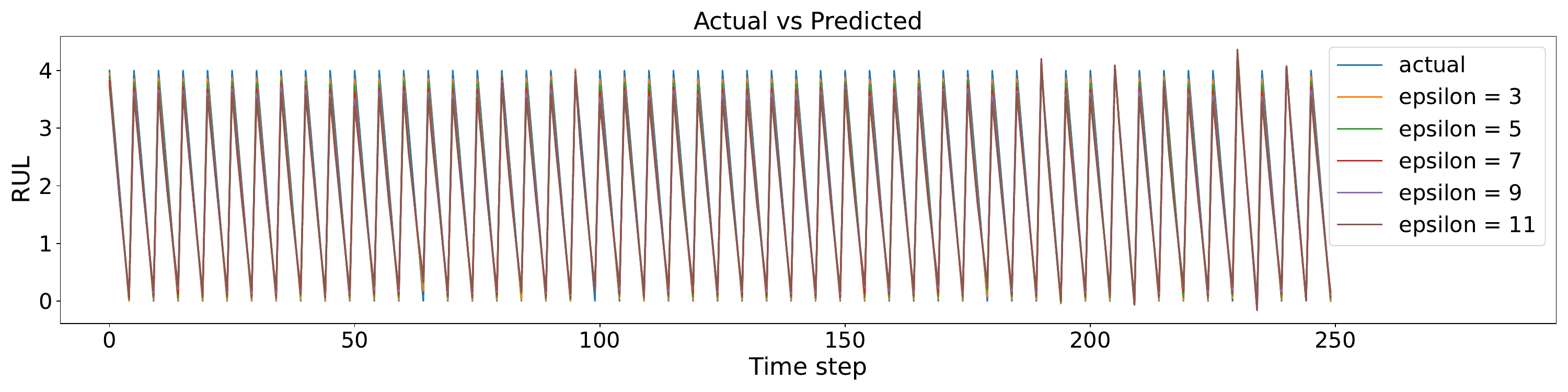}
  \caption{Predictions on the FGSM perturbed HDD test dataset for varying $\epsilon$.}
  \label{fig:hdd-fgsm-curve}
\end{figure}

\begin{figure}[!htb]
  \centering
  \includegraphics[width=\linewidth]{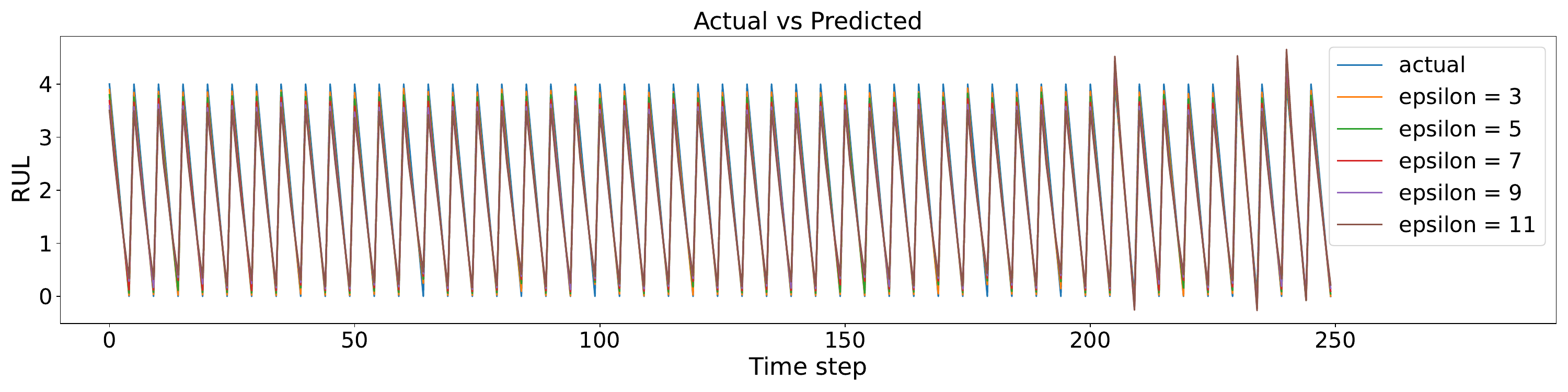}
  \caption{Predictions on the BIM perturbed HDD test dataset for varying $\epsilon$.}
  \label{fig:hdd-bim-curve}
\end{figure}

\begin{figure}[!htb]
  \centering
  \includegraphics[width=\linewidth]{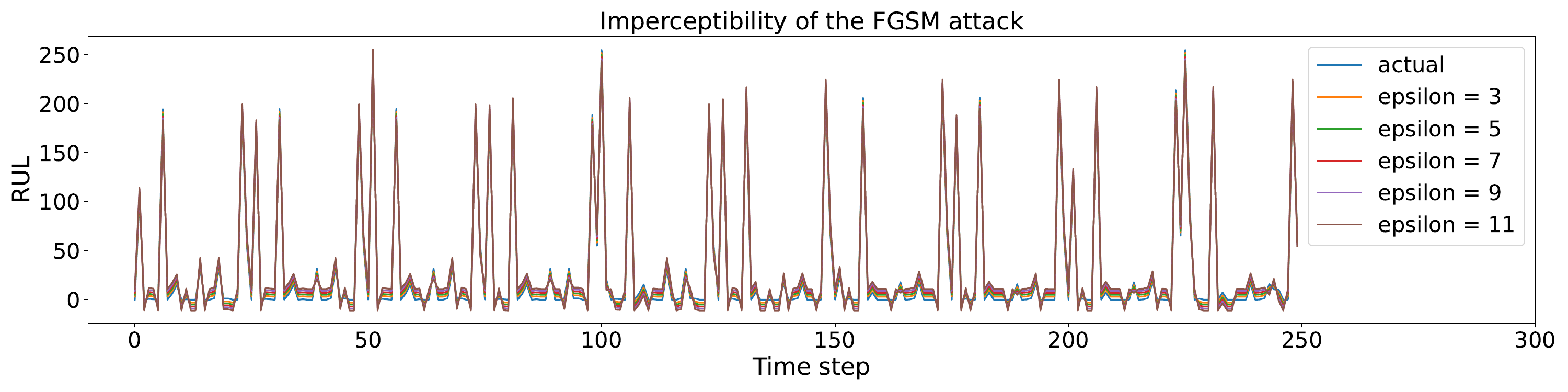}
  \caption{Imperceptibility of the FGSM attack on the HDD dataset.}
  \label{fig:hdd-fgsm-imperceptible}
\end{figure}

\begin{figure}[!htb]
  \centering
  \includegraphics[width=\linewidth]{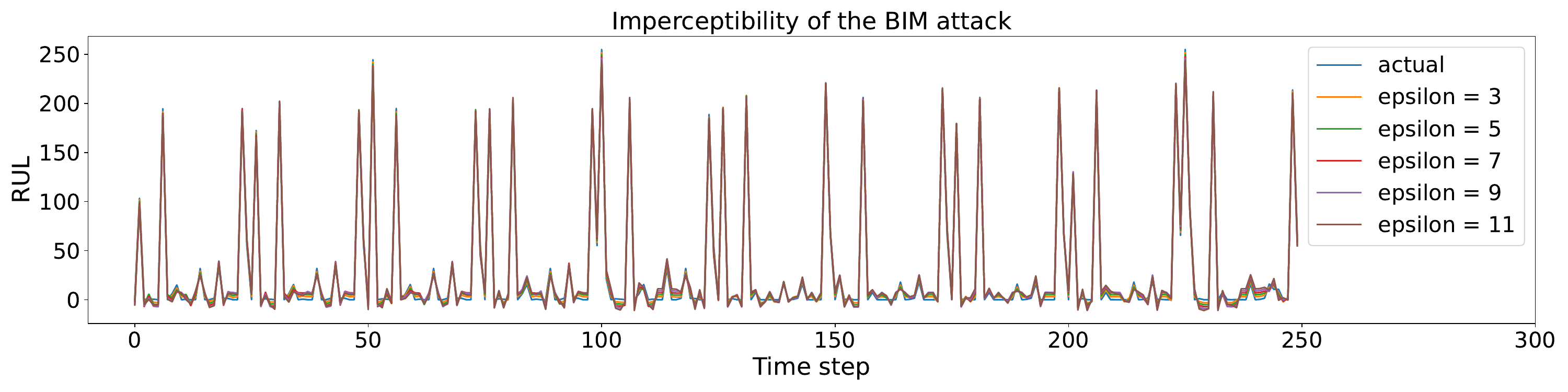}
  \caption{Imperceptibility of the BIM attack on the HDD dataset.}
  \label{fig:hdd-bim-imperceptible}
\end{figure}

We see that BIM is a stronger attack than FGSM in \figurename~\ref{fig:hdd-attack-rmse}.
\figurename~\ref{fig:hdd-fgsm-curve} and \figurename~\ref{fig:hdd-bim-curve} show the forecasts on the disrupted test set corresponding to the FGSM and BIM attacks respectively. \figurename~\ref{fig:hdd-fgsm-imperceptible} and \figurename~\ref{fig:hdd-bim-imperceptible} show the changes in a subset of input data after performing the FGSM and BIM attacks respectively thereby substantiating the fact that predictions on the attacked test set are worse while the inputs look the same to the naked eye.

\subsubsection{Results of Adversarial Defenses}
We performed two types of adversarial defenses to make the best-performing Encoder-Decoder LSTM more resilient to adversarial attacks seen in Section~\ref{sec:hdd-attack-results}. In the first method, we augmented the input feature samples with perturbed data in the course of model training to permit the model to learn the true values given perturbed features. The results of the different RMSE values obtained while performing DAAT on the HDD dataset are tabulated in \tablename~\ref{tbl:hdd-daat}. Although BIM is stronger than FGSM, DAAT on the HDD dataset is more resilient to the BIM attack due to the wide variations in feature values enabling the model to detect the noisy outliers in the underlying distribution better, as seen earlier.

\begin{table}[!htb] \small
\caption{Results after Performing DAAT on the HDD Dataset}
\label{tbl:hdd-daat}
\centering
\begin{tabular}{l|l|l}
\hline \hline
\textbf{Type of Attack} & \textbf{RMSE Metrics} & \textbf{Values} \\ \hline \hline
\multirow{4}{*}{FGSM} & Train & 0.0334 \\ \cline{2-3}
 & Validation & 0.0858 \\ \cline{2-3}
 & Test (clean data) & 0.0206 \\ \cline{2-3}
 & Test (poisoned data) & 0.0313 \\ \hline
\multirow{4}{*}{BIM} & Train & 0.0285 \\ \cline{2-3}
 & Validation & 0.0425 \\ \cline{2-3}
 & Test (clean data) & 0.0129 \\ \cline{2-3}
 & Test (poisoned data) & 0.0130 \\ \hline
\end{tabular}
\end{table}

In the next method, we perturb the gradients of the model during training hoping that the model learns a more robust distribution. We modified LPAT to include two different types of gradient perturbation during training - deterministic where $\epsilon$ is 7 i.e. the average of all expected $\epsilon$ values, and stochastic where $\epsilon$ takes a value between 3 and 11 i.e. the expected range. The RMSE metrics after applying LPAT to the HDD dataset are tabulated in \tablename~\ref{tbl:hdd-lpat}. 
\begin{table}[!htb] \small
\caption{Results after Performing LPAT on the HDD Dataset}
\label{tbl:hdd-lpat}
\centering
\begin{tabular}{l|l|l|l}
\hline \hline
    \textbf{Attack Type} & \textbf{Training Type} & \textbf{RMSE Metrics} & \textbf{Value} \\ \hline \hline
    \multirow{4}{*}{FGSM} & \multirow{3}{*}{Deterministic} & Train & 0.23 \\ \cline{3-4}
    & & Test & 0.1422 \\ \cline{2-4} 
    & \multirow{3}{*}{Stochastic} & Train & 0.5457 \\ \cline {3-4}
    & & Test & 0.1457 \\ \hline
    \multirow{4}{*}{BIM} & \multirow{3}{*}{Deterministic} & Train & 0.3538 \\ \cline{3-4}
    & & Test & 0.0926 \\ \cline{2-4} 
    & \multirow{3}{*}{Stochastic} & Train & 0.2848 \\ \cline{3-4}
    & & Test & 0.0455 \\ \hline
\end{tabular}
\end{table}

\begin{figure}[!htb]
  \centering

  \includegraphics[width=0.85\linewidth]{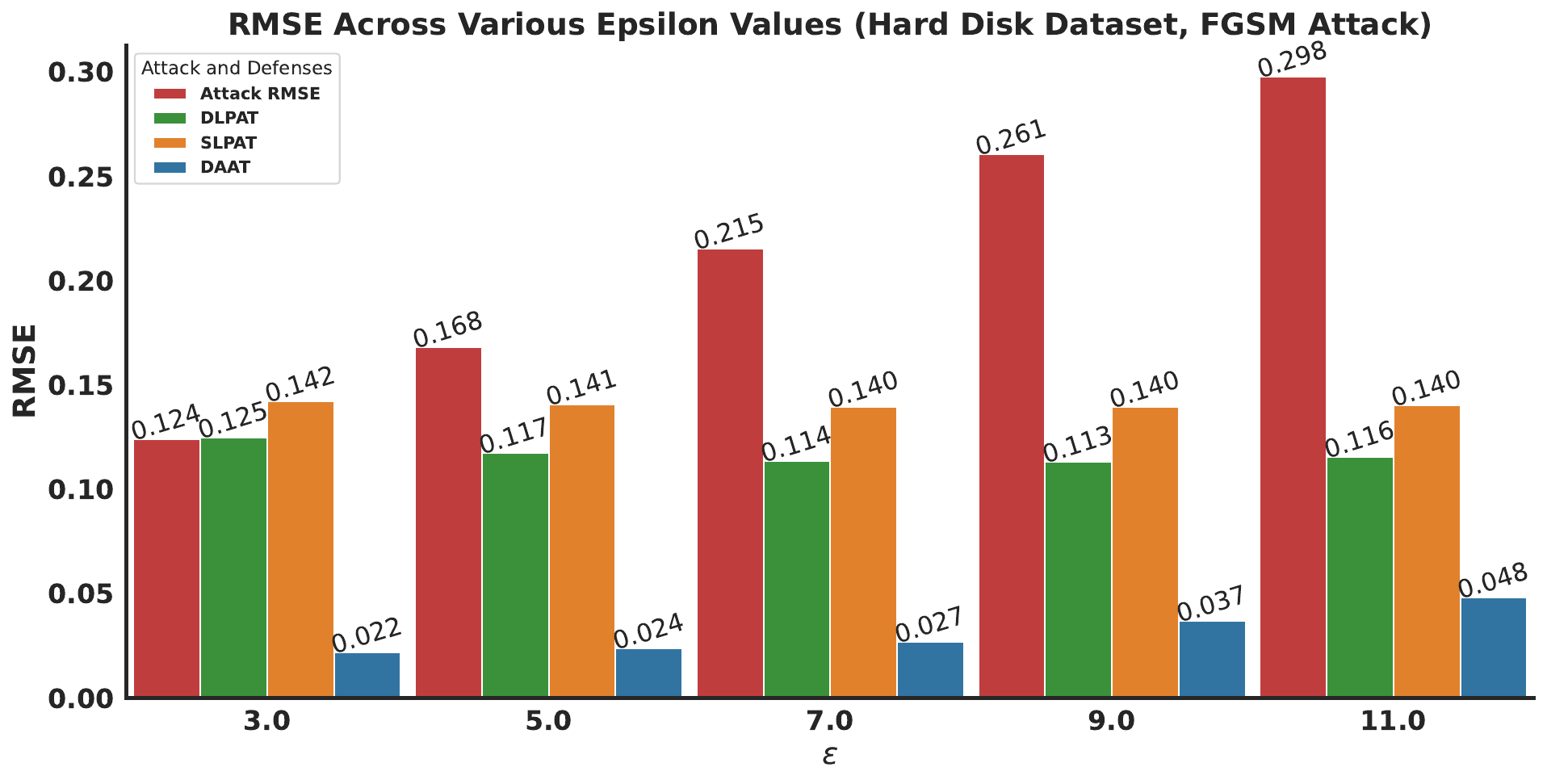}
  \caption{RMSE values of FGSM attack and defenses for HDD test set perturbed by varying $\epsilon$.}
  \label{fig:hdd-fgsm}
\end{figure}

\begin{figure}[!htb]
  \centering

  \includegraphics[width=0.85\linewidth]{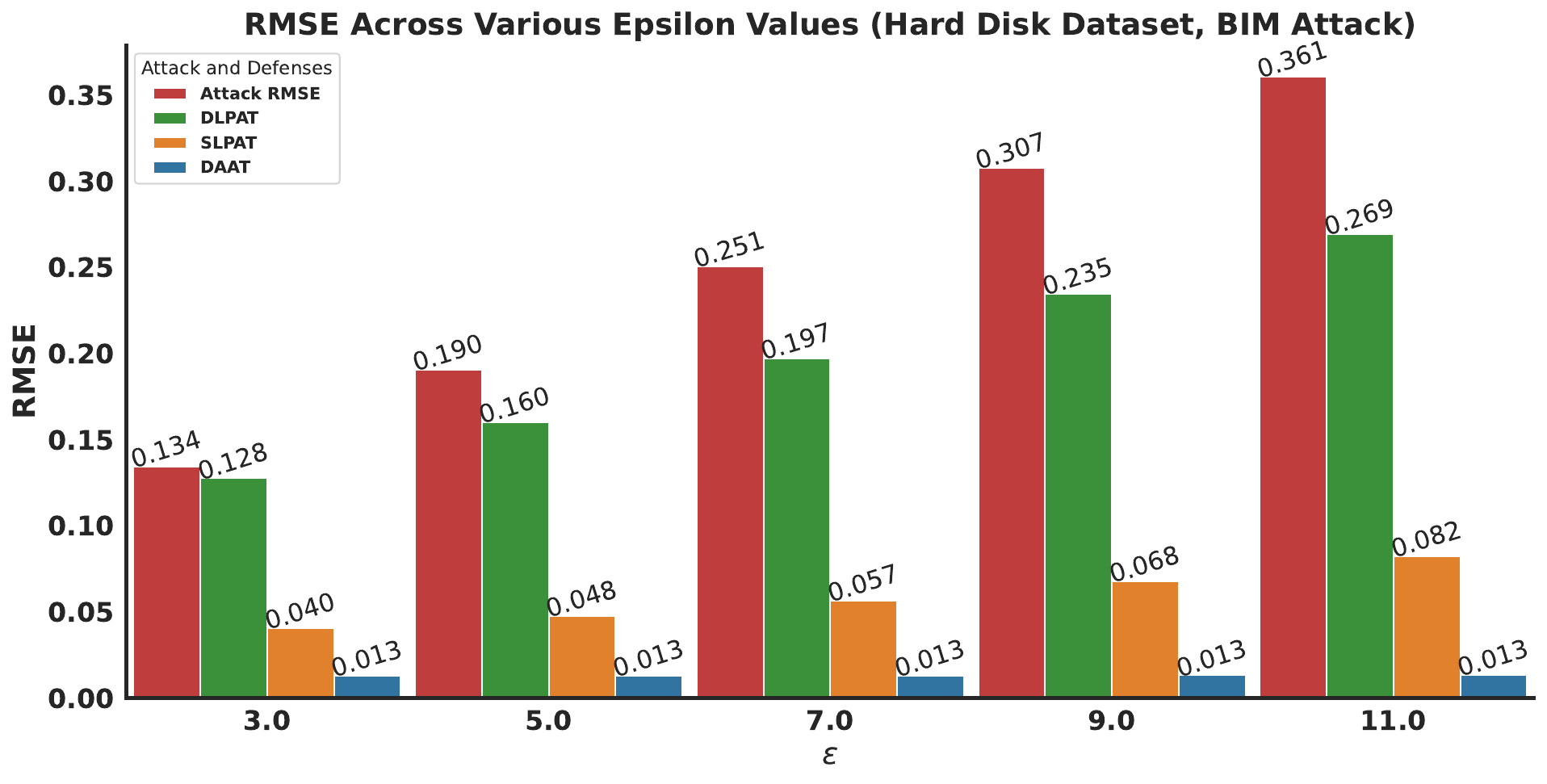}
  \caption{RMSE values of BIM attack and defenses for HDD test set perturbed by varying $\epsilon$.}
  \label{fig:hdd-bim}
\end{figure}

\begin{figure}[!htb]
  \centering
  
  \includegraphics[width=0.85\linewidth]{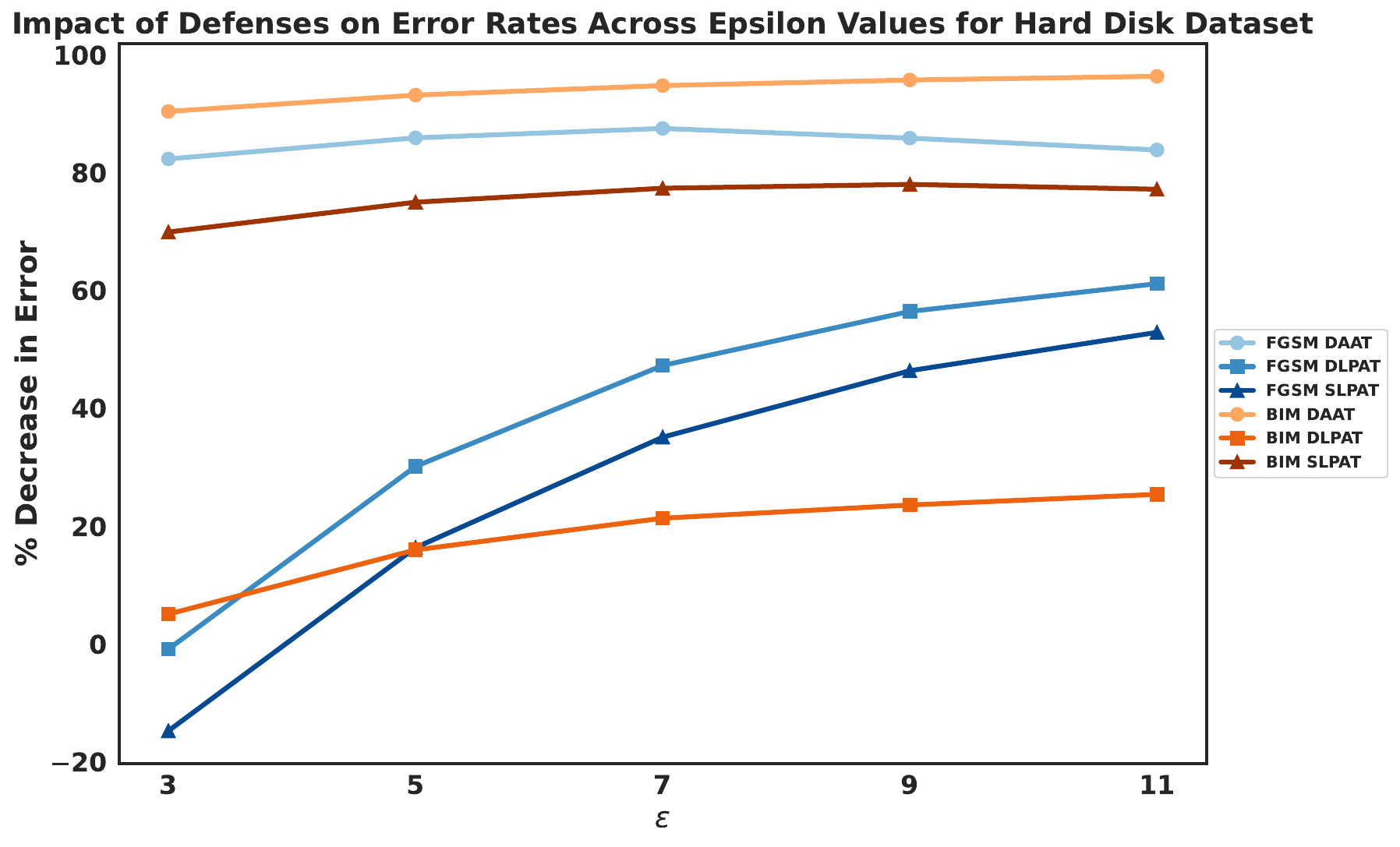}
  \caption{Percentage Reduction in Error Values after Adversarial Defense on the HDD Dataset}
  \label{fig:percent-decrease-hdd}
\end{figure}

\figurename~\ref{fig:hdd-fgsm} and \figurename~\ref{fig:hdd-bim} compare the RMSE values after the FGSM and BIM attacks and defenses respectively. \figurename~\ref{fig:percent-decrease-hdd} shows the reduction in error after defense for different $\epsilon$ and different attacks. 

It is seen that DLPAT and SLPAT for FGSM when $\epsilon$ = 3 have RMSE values higher than the attack RMSE. Although LPAT leads to more robust models resistant to attacks in most cases, in this case, we see that an average $\epsilon$ value of 7 did not help in learning a robust model when poisoned with a smaller $\epsilon$ value of 3. The same can be said for the deterministic gradient update where the random values picked during SLPAT do not effectively contribute to learning the underlying distribution with allowance for noise. 


\section{Limitations and Future Work}\label{sec:future_work}
Research in adversarial attacks and defenses for multivariate time-series forecasting is nascent, with many areas unexplored. This study focuses on white-box attacks, which necessitate knowledge of model parameters and inputs. In the future, we will explore black-box attacks to construct an attack model using the estimates of the target model alone. We aim to develop a model that can discern the adversary’s $\epsilon$ range based on prediction deviations and identify the maximum perturbation beyond which defense fails. Adversarial training with BIM as the perturbation method yields superior results, suggesting further exploration of the impact of $\epsilon$ on training, as BIM is a stronger attack than FGSM.  We find that stochastic gradient update is highly dependent on the random values of $\epsilon$ picked during each iteration of backpropagation, warranting research into the selection of $\epsilon$ during the training process Investigating black-box transferability and integrating DAAT and LPAT into a hybrid approach could enhance model robustness against adversarial attacks.

\section{Conclusion}\label{sec:conclusion}
Our study investigates the susceptibility of deep learning models in multivariate time-series forecasting to adversarial attacks and evaluates defense mechanisms. We show that models like vanilla LSTM and Encoder-Decoder LSTM when tested on the Individual Household Power Consumption and Backblaze Hard Disk Drive datasets, undergo significant performance degradation under adversarial perturbations like FGSM and BIM. The average error rate increases by 248.17\% and 223\% on the electricity and HDD datasets respectively, highlighting the impact of these attacks. 

We also visualize the subtle changes to the input distribution post-attack, which are not easily detectable. To counter these vulnerabilities, we implement two robust defenses: Data Augmentation-based Adversarial Training (DAAT) and Layer-wise Perturbation-based Adversarial Training (LPAT). DAAT, particularly with BIM for augmentation, greatly enhances model resilience, reducing errors by up to 72.41\% and 94.81\% for the electricity and HDD datasets respectively. 

LPAT also shows effectiveness, with performance varying based on perturbation magnitude. These results emphasize the need for adversarial defense strategies in deep learning models for critical applications in smart and connected infrastructures, enhancing model reliability and ensuring secure, dependable predictions.

\section*{Acknowledgment}
The research reported in this publication was supported by the Division of Research and Innovation at San Jos\'e State University under Award Number 23-UGA-08-044. The content is solely the responsibility of the author(s) and does not necessarily represent the official views of San Jos\'e State University. The authors would like to thank Dr. Sanchari Das for her proofreading and insightful feedback.

\raggedright
\bibliographystyle{apacite}
\bibliography{phm}

\justifying
\appendix
\section*{Appendix}
\section{Time-Series Data Characteristics}
\begin{figure}[!htb]
    \centering
    \begin{subfigure}[b]{0.45\linewidth}
        \includegraphics[width=\linewidth]{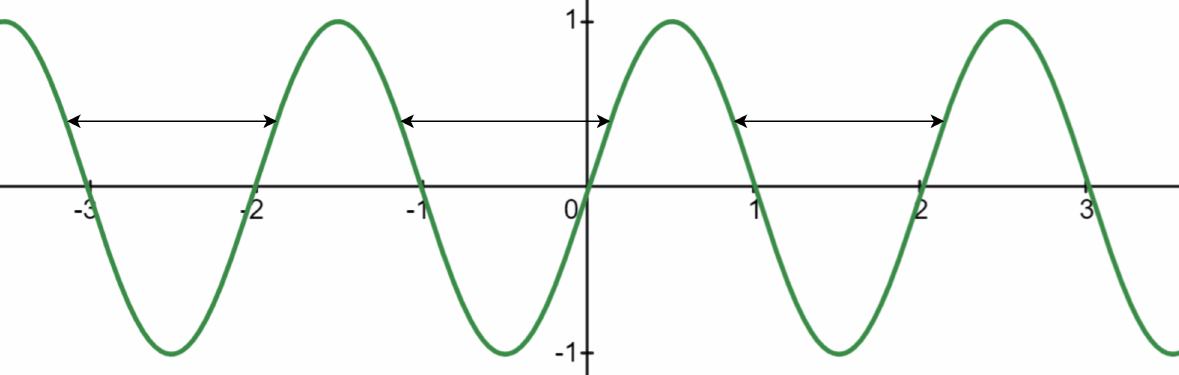}
        \caption{Stationary time-series}
        \label{fig:subfig1}
    \end{subfigure}
    \\
    \begin{subfigure}[b]{0.45\linewidth}
        \includegraphics[width=\linewidth]{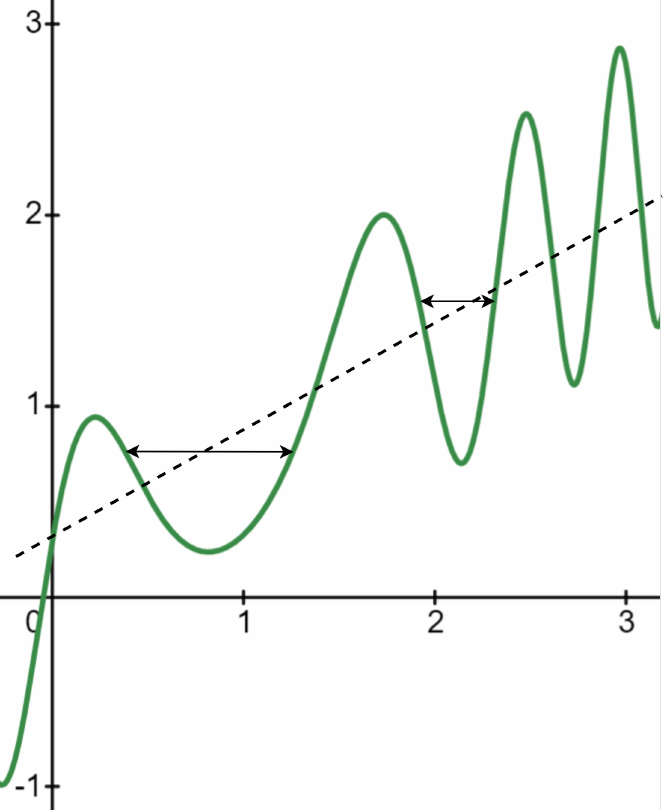}
        \caption{Non-stationary time-series}
        \label{fig:subfig2}
    \end{subfigure}
    \caption{Characteristics of time-series data}
    \label{fig:time-series}
\end{figure}

Time-series data are characterized into two types - stationary time-series where the mean or variance is a constant given by $\mu = c$ or $\sigma = c$ or non-stationary time-series in which the mean or variance varies with time given by $\mu = f(t)$ or $\sigma = f(t)$. \figurename~\ref{fig:time-series} shows a stationary time-series with constant mean and variance, and a non-stationary time-series where the mean increases with time as shown by the dotted line, and the variance representing the distance between consecutive peaks decreases with time. 

Time-series data can be represented as a 2D array as:
\begin{equation*}
\begin{bmatrix}
    v_{11} & v_{12} & \dots & v_{1n} \\
    v_{21} & v_{22} & \dots & v_{2n} \\
    \vdots & \vdots & \ddots & \vdots \\
    v_{t1} & v_{t2} & \dots & v_{tn}
\end{bmatrix}
\end{equation*}
where \textit{n} is the total readings recorded at any particular period, and \textit{t} is the total readings recorded over \textit{t} time steps. There are two types of time-series. The first is univariate time-series in which the future values of the series are dependent only on its past values. For example, in univariate time-series, the value of \(v_{tn}\) depends only on its past values such as \(\left[ v_{1n}, \, v_{2n}, \, \dots, \, v_{(t-1)n} \right]\). The second is a multivariate time-series in which the future values depend on a combination of the past values and predictors such as exogenous variables other than the series. For example, in multivariate time-series, the value of \(v_{tn}\) depends on both its past values and other parameters captured by sensing devices such as $[ v_{11}, \, v_{12}, \, \dots, v_{1n}, \, v_{21}, \, v_{22}, \, \dots, v_{2n}, \, \dots, \, v_{t1}, \, v_{t2}, \dots, \\ \, v_{(t-1)n} ]$.

\section{Supplemental Background}
\cite{sparse-attack} propose a new black-box attack to apply sparse perturbations to image pixels leading to unnoticeable changes in the resultant image. \cite{momentum-based-attack} iteratively attacked the samples by introducing a momentum term that prevents the gradients from being stuck in a local maxima resulting in a much more powerful attack. A paper published by \cite{rotation-attack}, outlines the vulnerability of CNNs \cite{cnn} to benign rotations and transformations. \cite{min-perturb} use a log barrier-based optimization technique to solve the constrained optimization problem that aims to minimize the perturbation magnitude in adversarial attacks. \cite{transfer-attack} show that an intermediate-level attack ensures high transferability of adversarial attacks between architectures. \cite{universal-perturbations} identify the presence of global perturbations which are independent of the images and depend on the geometric modeling of the decision edges of deep learning algorithms. \cite{one-pixel-attack} proved that CNNs are susceptible to attacks of lower dimensions by modifying only one pixel based on differential evolution to perturb images.

\cite{sankaranarayanan} suggest efficient layer-wise training to prevent overfitting in deep networks. \cite{polytope-separation} propose a convex polytope-based separation of features during learning, such that independent variables of various targets are maximally separated from one another. \cite{bat} introduce a bilateral adversarial training framework by perturbing the labels and the features during the training process. \cite{optimization-training} perform adversarial training through robust optimization, exploring the universal transferability during training. \cite{learn2perturb} introduce a framework that introduces perturbations during the training and inference and efficiently learns to detect noise in the input. \cite{min-max-training} outline a process called adversarial distribution training in which the internal maximization function seeks to learn the worst distribution possible, and the external minimization function seeks to minimize the loss over the distribution. \cite{vulnerability-supression-loss} put forth a novel loss function called vulnerability suppression loss that aims to minimize the latent space feature distortion. \cite{gans} develop an iterative stochastic generator to generate diverse adversarial examples capable of exposing the vulnerabilities in the target model. \cite{liu} propose an attack scheme that introduces imperceptible perturbations to create poisoned examples and training mechanisms based on randomized smoothing to enhance model robustness. 

\section{Individual Household Power Consumption Dataset}
\subsection{Data Preprocessing}
\begin{figure}[!htb]
  \centering
  \includegraphics[width=0.7\linewidth]{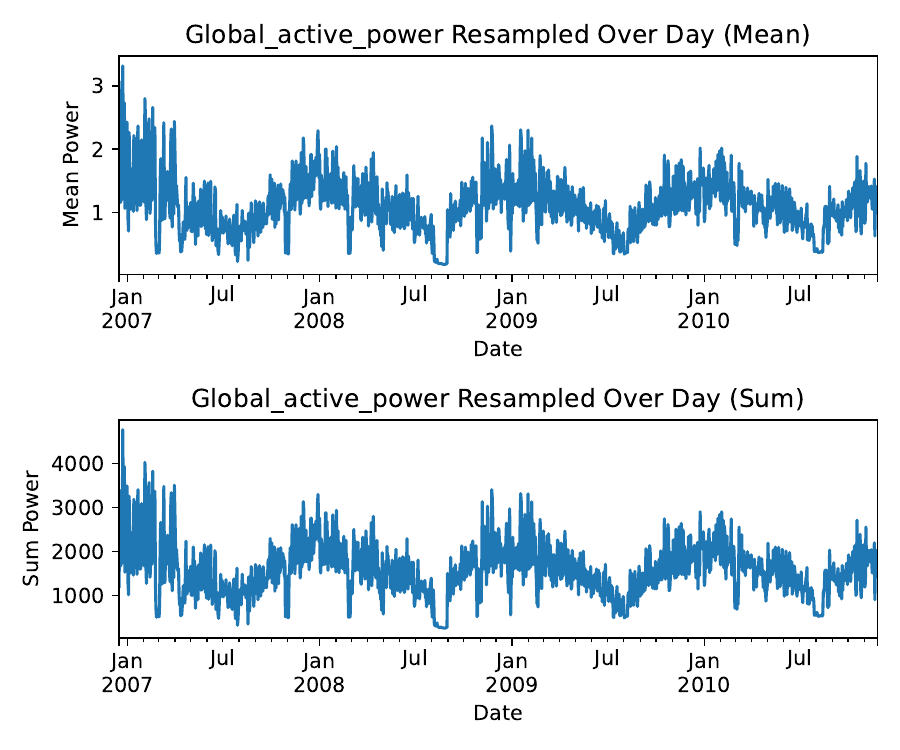}
  \caption{Global active power resampled per day for mean and sum of minutes.}
  \label{fig:mean_and_sum_of_mins}
\end{figure}

We conducted experiments using an LSTM which was designed to predict the \textit{global\_active\_power}. To facilitate this, we resampled the dataset daily, incorporating the mean of the per-minute values. \figurename~\ref{fig:mean_and_sum_of_mins} illustrates the distribution of the mean and sum of the per-minute values when the dataset is resampled daily. It is evident that whether we aggregate over the mean or sum while resampling over the day, the distribution remains consistent. Therefore, the pick of the clustering technique does not significantly impact the model estimates.
\vspace{-8pt}
\subsection{Architecture of vanilla LSTM}
\begin{figure}[!htb]
  \centering
  \includegraphics[width=0.7\linewidth]{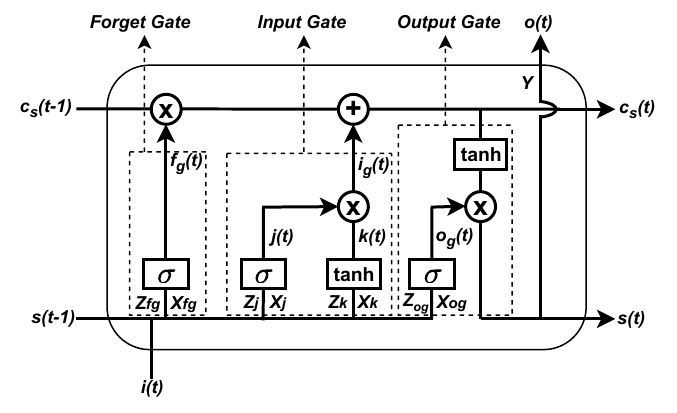}
  \caption{Vanilla LSTM unit}
  \label{fig:lstm}
\end{figure}

The vanilla LSTM unit shown in \figurename~\ref{fig:lstm} consists of:

\textit{1) Forget Gate:} The forget gate determines what piece of information to persist and what to delete. The input at a particular time step, i(t), and information from the preceding hidden state, s(t-1) act as inputs for the sigmoid activation, producing an output of 0 (forget), or 1 (remember). Its equation is:
\begin{equation}
f_g(t) = \sigma (i(t)X_f + s(t-1)Z_f)   
\end{equation}
\textit{2) Input Gate:} This gate decides the relevant information to be passed further from the present. The input vector and the previous hidden layer’s output are multiplied element by element, after passing through a sigmoid and a tanh function:
\begin{equation}
j(t) = \sigma(i(t)X_j + s(t-1)Z_{ig})
\end{equation}
\begin{equation}
k(t) = \tanh(i(t)X_k + s(t-1)Z_k)
\end{equation}
\begin{equation}
i_g(t) = j(t) \cdot k(t)
\end{equation}
\textit{3) Cell State:}It reserves relevant long-term memory and performs element-by-element multiplication of the ouput of the forget gate and previous cell state to preserve only the relevant state of the network, and then performs element-by-element addition with the output of the input gate, given by:
\begin{equation}
C_s(t) = \sigma(f_g(t) \cdot C_s(t - 1) + i_g(t))
\end{equation}
\textit{4) Output Gate:} This gate decides the resultant value at any period. The input vector, i(t), and the byproduct from the preceding hidden layer, s(t-1) are put into a sigmoid activation to calculate $o_g(t)$, which is then multiplied with the tanh of the new cell state $C_s(t)$, to pass on as intake of the next time step along with the cell state, $C_s(t)$. 
\begin{equation}
o_g(t) = \sigma(i(t)X_{og} + s(t - 1)Z_{og})
\end{equation}
\begin{equation}
s(t) = \tanh(C_s(t)) \cdot o_g(t)
\end{equation}

\subsection{Training Process}
\begin{figure}[!htb]
  \centering
  \includegraphics[width=0.7\linewidth]{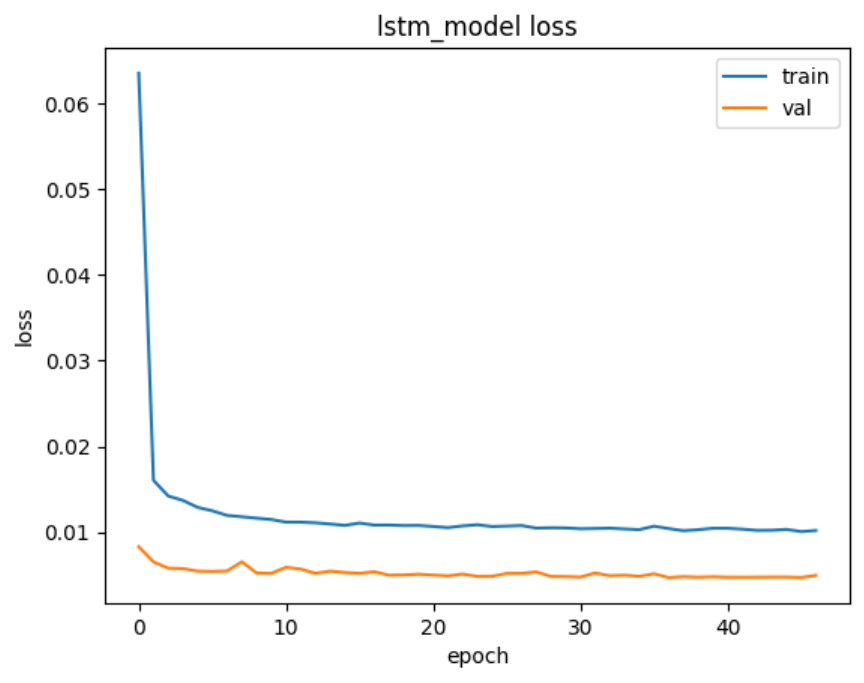}
  \caption{Loss function of the vanilla LSTM network.}
  \label{fig:loss_curve}
\end{figure}

Loss curve of the train and validation sets for the LSTM without using cross-validation or look back is shown in \figurename~\ref{fig:loss_curve}.

\begin{figure}[!htb]
  \centering
  \includegraphics[width=\linewidth]{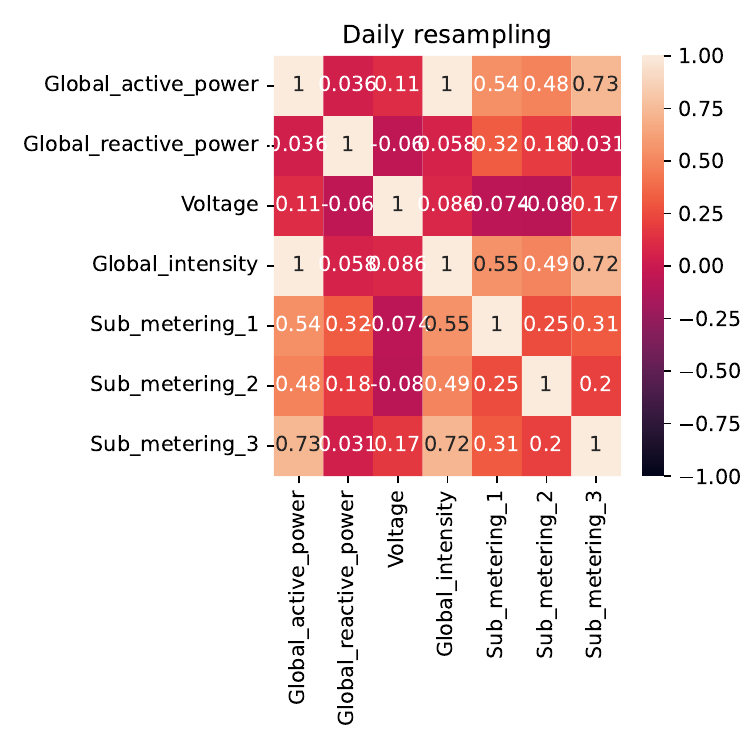}
  \caption{Correlation matrix of values resampled per day.}
  \label{fig:correlation_matrix_daily}
\end{figure}

\figurename~\ref{fig:correlation_matrix_daily} shows the correlation between features for the data resampled over days. Since each feature is highly correlated with itself, the diagonal values of the correlation matrix are 1. 

We perform feature selection based on the correlation matrix to check if it improves results. We identified the top features as \textit{global active power, global intensity, sub-metering 1, and sub-metering 3}. Training the LSTM using these features and without using cross-validation or look-back gives us a train RMSE of 0.1024 and a test RMSE of 0.0783 compared to the test RMSE of 0.0807 without using cross-validation thereby improving model efficacy by 2.97\%. 

\subsection{Adversarial Defense Results}
The overall percentage decrease in error on the electricity dataset after performing the different fortifications is tabulated in \tablename~\ref{tbl:perecent-decrease-elec} and is illustrated in \figurename~\ref{fig:percent-decrease-elec}.

\begin{table}[!htb] \small
\caption{Percentage Reduction in Error Values after Adversarial Training on the Electricity Dataset}
\label{tbl:perecent-decrease-elec}
\centering
\begin{tabular}{l|l|l|l}
\hline \hline
\textbf{Attack Type} & \textbf{Training Type} & \textbf{$\epsilon$} & \textbf{\% decrease in error} \\
\hline \hline
    \multirow{15}{*}{FGSM} & \multirow{5}{*}{DAAT} & 0.05 & 33.08 \\ \cline{3-4}
    & & 0.1  & 52.04 \\ \cline{3-4}
    & & 0.15 & 63.38 \\ \cline{3-4}
    & & 0.2  & 70.62 \\ \cline{3-4}
    & & 0.25 & 75.17 \\ \cline{2-4}
    & \multirow{5}{*}{DLPAT} & 0.05 & 16.67 \\ \cline{3-4}
    & & 0.1  & 43.94 \\ \cline{3-4}
    & & 0.15 & 58.73 \\ \cline{3-4}
    & & 0.2  & 67.72 \\ \cline{3-4}
    & & 0.25 & 73.71 \\ \cline{2-4}
    & \multirow{5}{*}{SLPAT} & 0.05 & 15.47 \\ \cline{3-4}
    & & 0.1  & 42.56 \\ \cline{3-4}
    & & 0.15 & 57.22 \\ \cline{3-4}
    & & 0.2  & 66.14 \\ \cline{3-4}
    & & 0.25 & 72.08 \\ \hline
    \multirow{15}{*}{BIM}  & \multirow{5}{*}{DAAT} & 0.05 & 32.79 \\ \cline{3-4}
    & & 0.1  & 54.66 \\ \cline{3-4}
    & & 0.15 & 66.02 \\ \cline{3-4}
    & & 0.2  & 72.7 \\ \cline{3-4}
    & & 0.25 & 76.55 \\ \cline{2-4}
    & \multirow{5}{*}{DLPAT} & 0.05 & 21.83 \\ \cline{3-4}
    & & 0.1  & 47.95 \\ \cline{3-4}
    & & 0.15 & 60.54 \\ \cline{3-4}
    & & 0.2  & 67.56 \\ \cline{3-4}
    & & 0.25 & 71.15 \\ \cline{2-4}
    & \multirow{5}{*}{SLPAT} & 0.05 & 19.81 \\ \cline{3-4}
    & & 0.1  & 47.41 \\ \cline{3-4}
    & & 0.15 & 61.17 \\ \cline{3-4}
    & & 0.2  & 69.09 \\ \cline{3-4}
    & & 0.25 & 73.38 \\ \hline
\end{tabular}
\end{table}

\section{Backblaze Hard Disk Drive Dataset}
\subsection{Architecture of Encoder-Decoder LSTM}
\begin{figure}[!htb]
  \centering
  \includegraphics[width=0.7\linewidth]{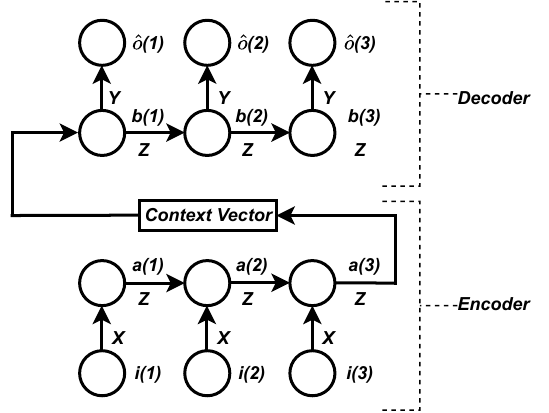}
  \caption{Architecture of Encoder Decoder LSTM.}
  \label{fig:encoder-decoder}
\end{figure}

The components of the Encoder-Decoder model used on this dataset are explained below and illustrated in \figurename~\ref{fig:encoder-decoder}.

\textit{1) Encoder:}
The encoder is composed of several RNNs, LSTMs, or GRUs stacked together accepting an input and propagating that information forward to the next units. The hidden state a(t) is computed from the input array, i(t), and the byproduct of the previous layer, a(t-1) based on the mapping of the chosen unit, if it is RNN, LSTM, or GRU. The final a(t) is composed of all the encoded information from the previous states and hidden layers. Its equation is given by:
\begin{equation}
a(t) = f(Z a(t-1) + X i(t))
\end{equation}

\textit{2) Context Vector:}
The context vector represents the last hidden state output of the encoder and the first hidden state input to the decoder. It is an encoded latent space representation of the inputs and allows the decoder to forecast accurately.

\textit{3) Decoder:}
The decoder consists of similar stacked recurrent RNN, LSTM, or GRU units, receiving a hidden state from the preceding unit, and calculates its hidden state and the output.
\begin{equation}
b(t) = f(Z a(t - 1))
\end{equation}
Softmax activation is applied on the hidden state, b(t), and the corresponding weight to output a probability vector as:
\begin{equation}
\hat{o}(t) = \text{softmax}(Y b(t))
\end{equation}

\subsection{Training Process}
\begin{figure}[!htb]
  \centering
  \includegraphics[width=0.7\linewidth]{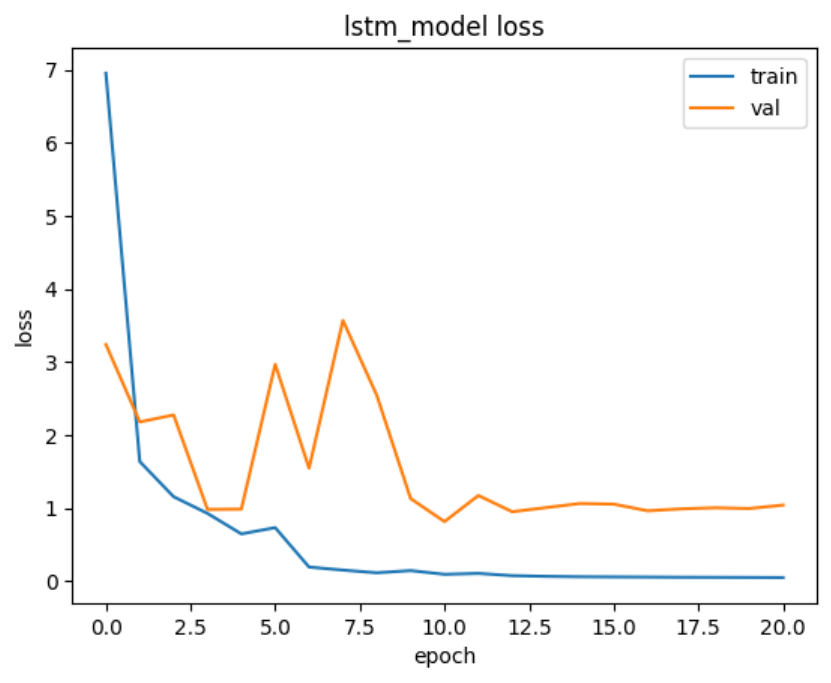}
  \caption{Loss function of Encoder Decoder LSTM.}
  \label{fig:hdd-loss}
\end{figure}

The loss function of the 5-day look back Encoder-Decoder model without cross-validation is shown in \figurename~\ref{fig:hdd-loss}.

We perform feature selection by training the 25-day look-back Encoder-Decoder LSTM with the top 10 features identified by an eXtreme Gradient Boosting (XGB) classifier as reported by \cite{hdd-rohan}. We found the train RMSE to be 12.1174 and the test RMSE to be 16.5570 which is much higher than those reported in \tablename~\ref{tbl:hdd-results}. 

\subsection{Adversarial Defense Results}
\tablename~\ref{tbl:percent-decrease-hdd} shows the reduction in error after performing adversarial defense for different perturbation values and different adversarial attacks. and is visualized in \figurename~\ref{fig:percent-decrease-hdd}.
\vspace*{-200pt}

\newpage

\begin{table}[!htb] \small
\caption{Percentage Reduction in Error Values after Adversarial Training on the HDD Dataset}
\label{tbl:percent-decrease-hdd}
\centering
\begin{tabular}{l|l|l|l}
\hline \hline
\textbf{Attack Type} & \textbf{Training Type} & \textbf{$\epsilon$} & \textbf{\% decrease in error} \\
\hline \hline
    \multirow{15}{*}{FGSM} & \multirow{5}{*}{DAAT} & 3 & 82.34 \\ \cline{3-4}
    & & 5 & 85.92 \\ \cline{3-4}
    & & 7 & 87.51 \\ \cline{3-4}
    & & 9 & 85.87 \\ \cline{3-4}
    & & 11 & 83.86 \\ \cline{2-4}
    & \multirow{5}{*}{DLPAT} & 3 & -0.81 \\ \cline{3-4}
    & & 5  & 30.18 \\ \cline{3-4}
    & & 7 & 47.31 \\ \cline{3-4}
    & & 9  & 56.47 \\ \cline{3-4}
    & & 11 & 61.18 \\ \cline{2-4}
    & \multirow{5}{*}{SLPAT} & 3 & -14.68 \\ \cline{3-4}
    & & 5 & 16.4 \\ \cline{3-4}
    & & 7 & 35.14 \\ \cline{3-4}
    & & 9 & 46.41 \\ \cline{3-4}
    & & 11 & 52.92 \\ \hline
    \multirow{15}{*}{BIM} & \multirow{5}{*}{DAAT} & 3 & 90.4 \\ \cline{3-4}
    & & 5 & 93.17 \\ \cline{3-4}
    & & 7 & 94.78 \\ \cline{3-4}
    & & 9 & 95.73 \\ \cline{3-4}
    & & 11 & 96.35 \\ \cline{2-4}
    & \multirow{5}{*}{DLPAT} & 3 & 5.13 \\ \cline{3-4}
    & & 5  & 16.02 \\ \cline{3-4}
    & & 7 & 21.4 \\ \cline{3-4}
    & & 9  & 23.64 \\ \cline{3-4}
    & & 11 & 25.44 \\ \cline{2-4}
    & \multirow{5}{*}{SLPAT} & 3 & 69.94 \\ \cline{3-4}
    & & 5 & 75 \\ \cline{3-4}
    & & 7 & 77.37 \\ \cline{3-4}
    & & 9 & 78.02 \\ \cline{3-4}
    & & 11 & 77.2 \\ \hline
\end{tabular}
\end{table}

\end{document}